\documentclass[journal]{IEEEtran}

\newcommand{\gmo}[0]{{GMO}}
\newcommand{\gso}[0]{{GSO}}
\newcommand{\name}[0]{{Cam4DOcc}}
\newcommand{\netname}[0]{{OCFNet}}

\let \sss= \scriptscriptstyle

\usepackage{graphicx}
\usepackage{hyperref} 
\hypersetup{colorlinks=true,linkcolor=blue,anchorcolor=blue,citecolor=blue}
\usepackage{multirow}
\usepackage{setspace}
\usepackage{multicol}
\usepackage{stfloats}

\usepackage{graphics}           
\usepackage{times}              
\usepackage{amsmath}            
\usepackage{amssymb}            
\usepackage{graphicx}
\usepackage{algorithm}
\usepackage[noend]{algpseudocode}
\usepackage{booktabs}
\usepackage{color}
\definecolor{instructioncolor}{rgb}{.5,.5,.5}

\usepackage[font=small]{caption}

\def\secref#1{Sec.~\ref{#1}}
\def\figref#1{Fig.~\ref{#1}}
\def\tabref#1{Tab.~\ref{#1}}
\def\eqref#1{Eq.~(\ref{#1})}

\makeatletter
\usepackage{xspace}
\DeclareRobustCommand\onedot{\futurelet\@let@token\@onedot}
\def\@onedot{\ifx\@let@token.\else.\null\fi\xspace}

\def\etal{{et al}\onedot}
\makeatother

\usepackage{array}
\newcolumntype{L}[1]{>{\raggedright\let\newline\\\arraybackslash\hspace{0pt}}m{#1}}
\newcolumntype{C}[1]{>{\centering\let\newline\\\arraybackslash\hspace{0pt}}m{#1}}
\newcolumntype{R}[1]{>{\raggedleft\let\newline\\\arraybackslash\hspace{0pt}}m{#1}}

\ifCLASSINFOpdf
\else
\fi

\begin{document}

\title{\huge \name: Benchmark for Camera-Only 4D Occupancy Forecasting in Autonomous Driving Applications}%
%
%

\author{Junyi~Ma$^{\dag}$, Xieyuanli~Chen$^{\dag}$, Jiawei~Huang, Jingyi~Xu, Zhen~Luo, \\ Jintao~Xu, Weihao~Gu, Rui~Ai, Hesheng~Wang$^{*}$
\thanks{Junyi~Ma, Jingyi~Xu, and Hesheng~Wang are with Shanghai Jiao Tong University. Junyi~Ma is also with HAOMO.AI Technology Co., Ltd. Xieyuanli~Chen is with National University of Defense Technology. Jiawei~Huang, Jintao~Xu, Weihao~Gu, and Rui~Ai are with HAOMO.AI Technology Co., Ltd. Zhen~Luo is with Beijing Institute of Technology.}
\thanks{$^{\dag}$Equal contribution}
\thanks{$^{*}$Corresponding author email: wanghesheng@sjtu.edu.cn}
}

\maketitle

\begin{abstract}
Understanding how the surrounding environment changes is crucial for performing downstream tasks safely and reliably in autonomous driving applications. Recent occupancy estimation techniques using only camera images as input can provide dense occupancy representations of large-scale scenes based on the current observation. However, they are mostly limited to representing the current 3D space and do not consider the future state of surrounding objects along the time axis. To extend camera-only occupancy estimation into spatiotemporal prediction, we propose \name, a new benchmark for camera-only 4D occupancy forecasting, evaluating the surrounding scene changes in a near future. We build our benchmark based on multiple publicly available datasets, including nuScenes, nuScenes-Occupancy, and Lyft-Level5, which provides sequential occupancy states of general movable and static objects, as well as their 3D backward centripetal flow. To establish this benchmark for future research with comprehensive comparisons, we introduce four baseline types from diverse camera-based perception and prediction implementations, including a static-world occupancy model, voxelization of point cloud prediction, 2D-3D instance-based prediction, and our proposed novel end-to-end 4D occupancy forecasting network. Furthermore, the standardized evaluation protocol for preset multiple tasks is also provided to compare the performance of all the proposed baselines on present and future occupancy estimation with respect to objects of interest in autonomous driving scenarios. The dataset and our implementation of all four baselines in the proposed \name{} benchmark will be released here: \url{https://github.com/haomo-ai/Cam4DOcc}.
\end{abstract}


%
\IEEEpeerreviewmaketitle

\section{Introduction}
\label{sec:intro}

Accurately perceiving the status of objects in surrounding environments using cameras is important for autonomous vehicles or robots to make reasonable downstream planning and action decisions.
Traditional camera-based perception methods for object detection~\cite{liu2022petr,reading2021categorical,wang2023yolov7, zhou2022mogde}, semantic segmentation~\cite{xie2021segformer, Li_2022_CVPR, long2015fully, chen2017deeplab}, and panoptic segmentation~\cite{voedisch23codeps, hu2023you, li2023point2mask, cheng2020panoptic} focus on predefined specific object categories, making them less effective at recognizing uncommon objects. 
To tackle this limitation, a shift towards camera-based occupancy estimation~\cite{wang2023openoccupancy,huang2023tri,tong2023scene,li2023voxformer,wei2023surroundocc} has emerged by estimating the spatial occupancy states over classifying specific objects. It reduces the complexity of multi-class classification tasks and emphasizes general occupied state estimation, enhancing the reliability and adaptability of autonomous mobile systems.

\begin{figure}
  \centering
  \includegraphics[width=1\linewidth]{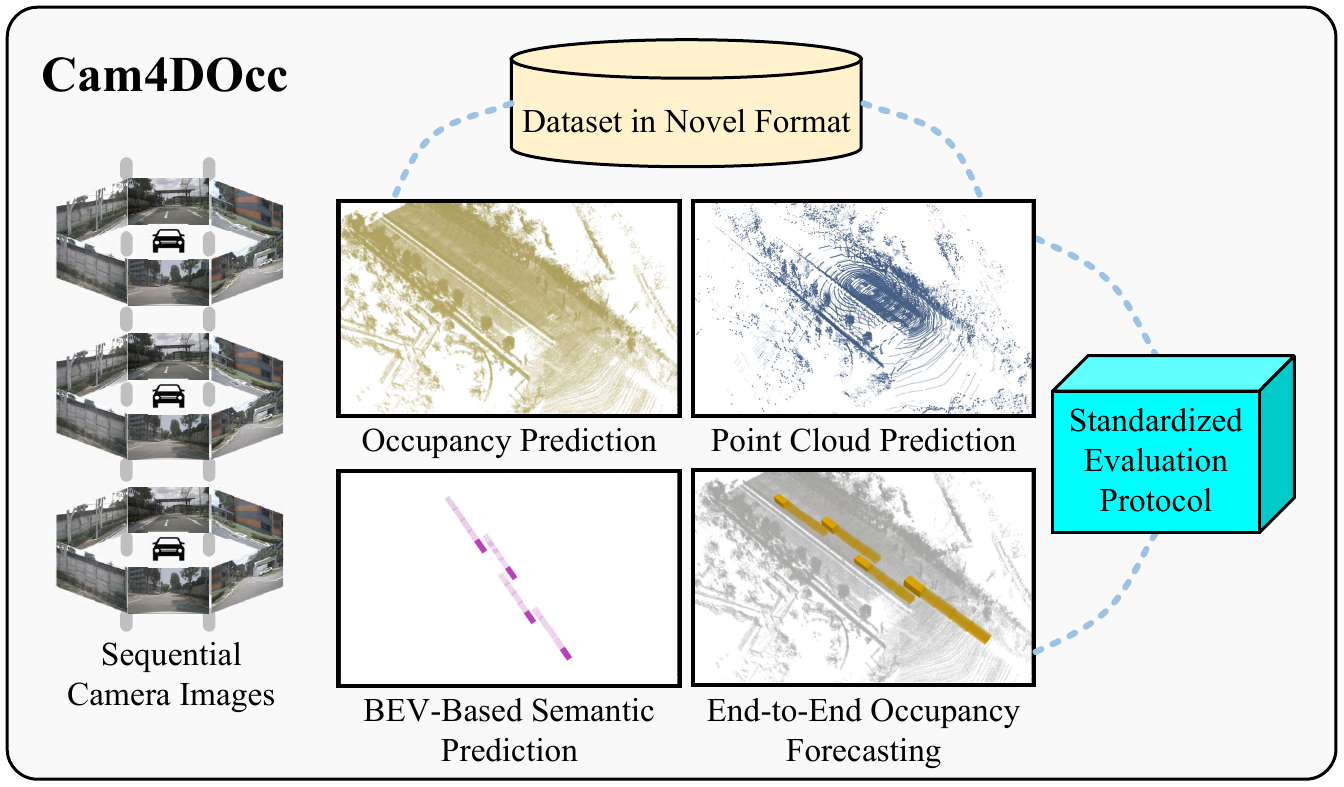}
  \caption{\name{} focuses on providing a novel dataset format, creating baselines modified from off-the-shelf camera-based perception and prediction approaches, and proposing a standardized evaluation protocol for the 4D occupancy forecasting task.}
  \label{fig:motivation}
  \vspace{-0.6cm}
\end{figure}

Despite the increasing attention to camera-based occupancy estimation, existing methods only estimate the current and past occupancy status. However, advanced collision avoidance and trajectory optimization methods employed by autonomous vehicles~\cite{ding2021epsilon,ding2019safe,song2020pip} require the ability to forecast future environmental conditions to ensure the safety and reliability of driving.
Some semantic/instance prediction algorithms~\cite{hu2021fiery, ijcai2023p120, wu2020motionnet, mahjourian2022occupancy, hendy2020fishing} have been proposed to forecast the motion of objects of interest, but they are mostly limited to 2D bird's eye view (BEV) format and can only recognize specific objects, mainly in the vehicle category. As to existing occupancy forecasting algorithms~\cite{khurana2023point, khurana2022differentiable,toyungyernsub2022dynamics} without considering semantics, they need LiDAR point clouds as necessary prior information to perceive the surrounding spatial structure, while LiDAR-based solutions are more resource-intensive and expensive than the camera counterparts.
It is natural to anticipate the next significant challenge in autonomous driving will be camera-only 4D occupancy forecasting.
This task aims to not only extend temporal occupancy prediction with camera images as input but also broaden semantic/instance prediction beyond BEV format and predefined categories. 
To this end, we propose \textit{\name} as shown in \figref{fig:motivation}, the first camera-only 4D occupancy forecasting benchmark comprising the new format of dataset, various types of baselines, and standardized evaluation protocol, to facilitate the advancements in this emerging domain. 
In this benchmark, we construct a dataset by extracting continuous occupancy changes along the time axis from the original nuScenes~\cite{caesar2020nuscenes}, nuScenes-Occupancy~\cite{wang2023openoccupancy}, and Lyft-Level5 \cite{lyft2019}. This dataset includes sequential semantic and instance annotations and 3D backward centripetal flow indicating the motion of occupancy grids.
Furthermore, to achieve camera-based 4D occupancy forecasting, we introduce four baseline methods, including a static-world occupancy model, voxelization of point cloud prediction, 2D-3D instance-based prediction, and an end-to-end 4D occupancy forecasting network. 
Finally, we evaluate the performance of these baseline methods for both present and future occupancy estimation using a proposed standardized protocol.

The main contributions of this paper are fourfold: (1) We propose \name{}, the first benchmark to facilitate future work on camera-based 4D occupancy forecasting. (2) We propose a new dataset format for the forecasting task in autonomous driving scenarios by leveraging existing datasets in the field. (3) We provide four novel baselines for camera-based 4D occupancy forecasting. Three of them are the extension of off-the-shelf approaches. Additionally, we introduce a novel end-to-end 4D occupancy forecasting network that demonstrates strong performance and can serve as a valuable reference for future research. (4) We introduce a novel standardized evaluation protocol and conduct comprehensive experiments with detailed analysis based on this protocol with our \name.


\section{Related Work}
\label{sec:related}

\textbf{Occupancy prediction.}~Occupancy prediction/estimation is a trendy technique to comprehensively estimate the occupancy state of the surrounding environments. It represents the space with geometric details significantly enhancing the expressiveness of complex scenes. 
MonoScene proposed by Cao~\etal~\cite{cao2022monoscene} first addresses 3D scene semantic completion from camera images, but only considers the front-view voxels. In contrast, Huang~\etal~\cite{huang2023tri} replace the Features Line of Sight Projection of MonoScene with TPVFormer to enhance the performance of surround-view occupancy prediction based on cross attention mechanism. UniOcc by Pan~\etal~\cite{pan2023uniocc} combines voxel-based neural radiance field (NeRF) with occupancy prediction to implement geometric and semantic rendering. Wang~\etal~\cite{wang2023openoccupancy} propose a large-scale benchmark named OpenOccupancy which establishes the nuScenes-Occupancy dataset with high-resolution occupancy ground-truth, and further provides several baselines using different modalities. 
Tong~\etal~\cite{tong2023scene} also propose an occupancy prediction benchmark OpenOcc and exploit the occupancy estimated by their OccNet on various tasks, including semantic scene completion, 3D object detection, BEV segmentation, and motion planning. More recently, Occ3D~\cite{tian2023occ3d} utilizes occlusion reasoning and image-guided refinement to further improve the annotation quality. 
Similar to OpenOcc, SurroundOcc by Wei~\etal~\cite{wei2023surroundocc} also produces dense occupancy labels and uses spatial attention to reproject 2D camera features back to the 3D volumes.

\textbf{Occupancy forecasting.}~Occupancy forecasting is utilized to foresee how the surrounding occupancy changes in the near future beyond the present moment. 
Existing occupancy forecasting approaches~\cite{khurana2023point,khurana2022differentiable,toyungyernsub2022dynamics} mainly use LiDAR point clouds as input to capture the change of surrounding structures. For example, Khurana~\etal~\cite{khurana2023point} propose a differentiable raycasting method to forecast 2D occupancy states by pose-aligned LiDAR sweeps. More recently, they propose rendering future pseudo LiDAR points with estimated occupancy~\cite{khurana2023point}. Other Point cloud prediction methods~\cite{fan2019pointrnn,lu2021monet,mersch2022self,luo2023pcpnet} directly forecast the future laser points, which can be voxelized to future occupancy estimation. However, they still need sequential LiDAR point clouds and lose semantic consistency during prediction. 
In contrast to the above-mentioned LiDAR-based occupancy forecasting,
directly predicting future 3D occupancy with multiple semantic categories using only camera images in large-scale scenes remains challenging. Therefore, some camera-only semantic/instance prediction methods turn to forecast the motion of objects of interest, e.g., general vehicle classes on 2D BEV occupancy representation~\cite{hu2021fiery,akan2022stretchbev,zhang2022beverse,ijcai2023p120}.
For example, FIERY by Hu~\etal~\cite{hu2021fiery} directly extracts BEV features from multi-view 2D camera images and then combines a temporal convolution model and a recurrent network to estimate future instance distributions. After that, StretchBEV~\cite{akan2022stretchbev} and BEVerse~\cite{zhang2022beverse} are proposed for further enhancement on longer time horizons. Towards the over-supervision with redundant outputs, PowerBEV~\cite{ijcai2023p120} is recently proposed to improve the forecasting performance on accuracy and efficiency. 

The abovementioned methods cannot directly achieve the camera-only 4D occupancy forecasting task. In this work, we propose a novel benchmark on this topic where several baselines are created by converting the implementation of the existing state-of-the-art occupancy prediction, point cloud prediction, and BEV-based semantic/instance prediction algorithms. In addition, we develop a novel camera-based 4D occupancy forecasting network that can simultaneously forecast the future occupancy state of the general movable and static objects end-to-end. Standardized dataset format and evaluation protocol are also proposed to train and test all the baselines, which can further support future work in this literature.

\section{\name{} Benchmark}
\label{sec:benchmark}

\begin{figure*}
  \centering
  \includegraphics[width=1\linewidth]{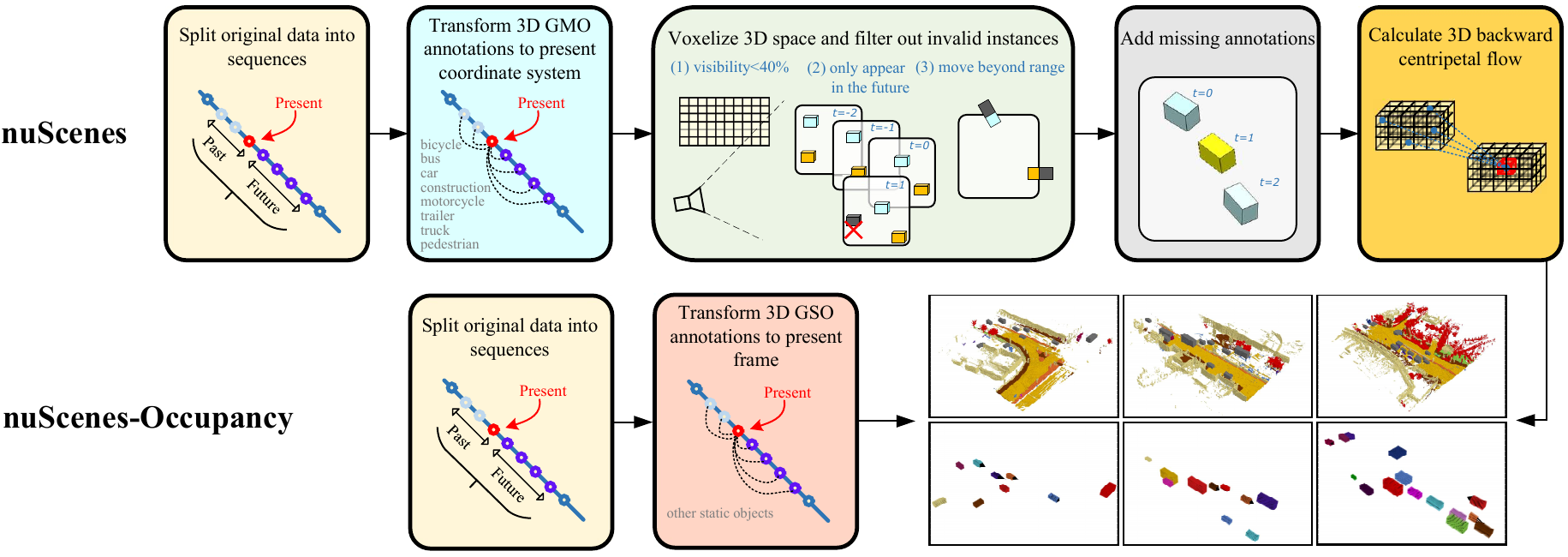}
  \vspace{-0.5cm}
  \caption{Overall pipeline of constructing dataset in our \name{} based on the original nuScenes and nuScenes-Occupancy. The dataset is reorganized into a novel format that considers both general movable and static categories for the unified 4D occupancy forecasting task.}
  \label{fig:dataset_building}
  \vspace{-0.5cm}
\end{figure*}

\subsection{Task Definition}

Given $N_p$ past and the current consecutive camera images $\mathcal{I}=\{I_t\}_{t=-N_p}^{0}$ as input, 4D occupancy forecasting aims to output the current occupancy $\mathbf{O}_c \in \mathbb{R}^{1 \times H \times W \times L}$ and the future occupancy $\mathbf{O}_f \in \mathbb{R}^{N_f \times H \times W \times L}$ in a short time interval $N_f$, where $H$, $W$, $L$ represent the height, width, and length of the specific range defined in the present coordinate system ($t=0$). Each voxel of $\mathbf{O}_f$ has $N_f$ sequential states $\mathcal{S}=\{S_t\}_{t=1}^{N_f}$ to represent whether it is free or occupied in each future timestamp.

\name{} considers two categories regarding their motion characteristics, \textit{general movable objects (\gmo)}, and \textit{general static objects (\gso)}, as the semantic labels of occupied voxel grids. \gmo{} usually have higher dynamic motion characteristics compared to \gso, thus requiring more attention during traffic activities for safety reasons. Accurately estimating the behavior of \gmo{} and predicting their potential motion changes significantly affect the decision making and motion planning of the ego vehicle. Compared to the previous semantic scene completion task~\cite{wang2023openoccupancy,huang2023tri,tong2023scene,chen20203d,li2020anisotropic,roldao2020lmscnet} considering multiple semantic categories, we focus more on investigating the ongoing change of voxel states for movable objects because we believe that motion characteristics of traffic participants deserve increased attention in the context of autonomous driving applications. Compared to the existing semantic/instance prediction task~\cite{hu2021fiery, ijcai2023p120, casas2021mp3,akan2022stretchbev,zhang2022beverse}, we not only emphasize the prediction of neighboring foreground objects but also focus on the occupancy estimation for the background of surrounding environments towards the requirement of more reliable navigation for autonomous vehicles.

\subsection{Dataset in New Format}
\label{sec:dataset_in_new_format}

Our \name{} benchmark introduces a new dataset format based on original nuScenes \cite{caesar2020nuscenes}, nuScenes-Occupancy \cite{wang2023openoccupancy}, and Lyft-Level5. As~\figref{fig:dataset_building} illustrates, we first split the original nuScenes dataset into sequences with the time length of $N=N_p+N_f+1$. Then sequential semantic and instance annotations of movable objects are extracted for each sequence and collected into the \gmo{} class, including \textit{bicycle}, \textit{bus}, \textit{car}, \textit{construction}, \textit{motorcycle}, \textit{trailer}, \textit{truck}, and \textit{pedestrian}. They are all transformed to the present coordinate system ($t=0$). After that, we voxelize the present 3D space and attach semantic/instance labels to the grids of movable objects using bounding boxes annotation. Notably, the invalid instance is discarded in this process once: (1) its visibility is under 40\% over the 6 camera images if it is a newly appeared object in $N_p$ historical frames, (2) it first appears in $N_f$ incoming frames, or (3) it moves beyond the range ($H, W, L$) predefined at $t=0$.
The visibility is quantified by the visible proportion of all pixels of the instance showing in camera images~\cite{caesar2020nuscenes}.
The sequential annotations are exploited to fill in missing intermediate instances based on constant velocity assumption~\cite{ijcai2023p120,chen2022auto}. The same operations are also applied to the Lyft-Level5 dataset. The distribution of instance duration $[t_{in}, t_{out}]$ after the processing mentioned above is presented in supplementary \secref{sec:dataset_setup_details}.
Lastly, we generate 3D backward centripetal flow using the instance association in the annotations. Li~\etal~\cite{ijcai2023p120} introduced 2D backward centripetal flow to improve the efficiency of 2D instance prediction. Inspired by that, we calculate 3D backward centripetal flow pointing from the voxel at time $t$ to its corresponding 3D instance center at $t-1$. We exploit this 3D flow to improve the accuracy of camera-based 4D occupancy forecasting (see \secref{sec:ab_on_flow}). 

We aim not only to forecast future positions of \gmo{} but also to estimate the occupancy state of \gso{} and free space necessary for safe navigation. Thus, we further concatenate the sequential instance annotations from the original nuScenes with the sequential occupancy annotations transformed to the present frame from nuScenes-Occupancy. This combination balances safety and precision for downstream navigation in autonomous driving applications. \gmo{} labels are borrowed from the bounding box annotations of the original nuScenes, which can be regarded as performing a dilation operation on the movable obstacles. \gso{} and free labels are provided by nuScenes-Occupancy to concentrate on more fine-grained geometric structures of surrounding large-scale environments. 

\subsection{Evaluation Protocol}
\label{sec:eval_proto}
To fully access the camera-only 4D occupancy forecasting performance, we establish various evaluation tasks and metrics with varying levels of complexity in our \name.

\textbf{Multiple tasks.} We introduce four-level occupancy forecasting tasks in the standardized evaluation protocol: (1) \textit{Forecasting inflated \gmo}: the categories of all the occupancy grids are divided into \gmo{} and others, where the voxel grids within the instance bounding boxes from nuScenes and Lyft-Level5 are annotated as \gmo{}. (2) \textit{Forecasting fine-grained \gmo}: the categories are also divided into \gmo{} and others but the annotation of \gmo{} are directly from voxel-wise labels of nuScenes-Occupancy removing invalid grids introduced in \secref{sec:dataset_in_new_format}. (3) \textit{Forecasting inflated \gmo{}, fine-grained \gso, and free space}: the categories are divided into \gmo{} from bounding box annotations, \gso{} following fine-grained annotations, and free space. (4) \textit{Forecasting fine-grained \gmo{}, fine-grained \gso, and free space}: the categories are divided into \gmo{} and \gso{} both following fine-grained annotations, and free space.
Since the Lyft-Level5 dataset lacks occupancy labels, we only conduct the evaluation for the first task on it.

\noindent\textbf{Metrics.} For all four tasks, we use intersection over union (IoU) as the performance metric. We separately evaluate the current moment ($t=0$) occupancy estimation and the future time ($t \in [1, N_f]$) forecasting by
\begin{small}
\begin{align}
\text{IoU}_c(\hat{\mathbf{O}}_c, \mathbf{O}_c) &= \frac{\sum_{\sss{H,W,L}}\hat{S}_c \cdot S_c}{\sum_{\sss{H,W,L}}\hat{S}_c+S_c-\hat{S}_c \cdot S_c},
\end{align}
\end{small}%
\begin{small}
\begin{align}
\text{IoU}_f(\hat{\mathbf{O}}_{f}, \mathbf{O}_{f}) &= \frac{1}{N_f}\sum_{t=1}^{N_f} \frac{\sum_{\sss{H,W,L}}\hat{S}_t \cdot S_t}{\sum_{\sss{H,W,L}}\hat{S}_t+S_t-\hat{S}_t \cdot S_t},
\label{eq:iou}
\end{align}
\end{small}%
where $\hat{S}_t$ and $S_t$ represent the estimated and ground-truth voxel state at timestamp $t$ respectively. 

We also provide a singular quantitative indicator to evaluate forecasting performance within the whole time horizon using one value calculated by
\begin{small}
\begin{align}
\tilde{\text{IoU}}_f(\hat{\mathbf{O}}_{f}, \mathbf{O}_{f}) &= \frac{1}{N_f}\sum_{t=1}^{N_f}\frac{1}{t}\sum_{k=1}^{t} \frac{\sum_{\sss{H,W,L}}\hat{S}_k \cdot S_k}{\sum_{\sss{H,W,L}}\hat{S}_k+S_k-\hat{S}_k \cdot S_k}.
\label{eq:iou_new}
\end{align}
\end{small}%
IoU of timestamps closer to the current moment contributes more to the final $\tilde{\text{IoU}}_f$. This aligns with the principle that occupancy predictions at near timestamps are more crucial for subsequent motion planning and decision making.

\begin{figure*}
  \centering
  \includegraphics[width=1\linewidth]{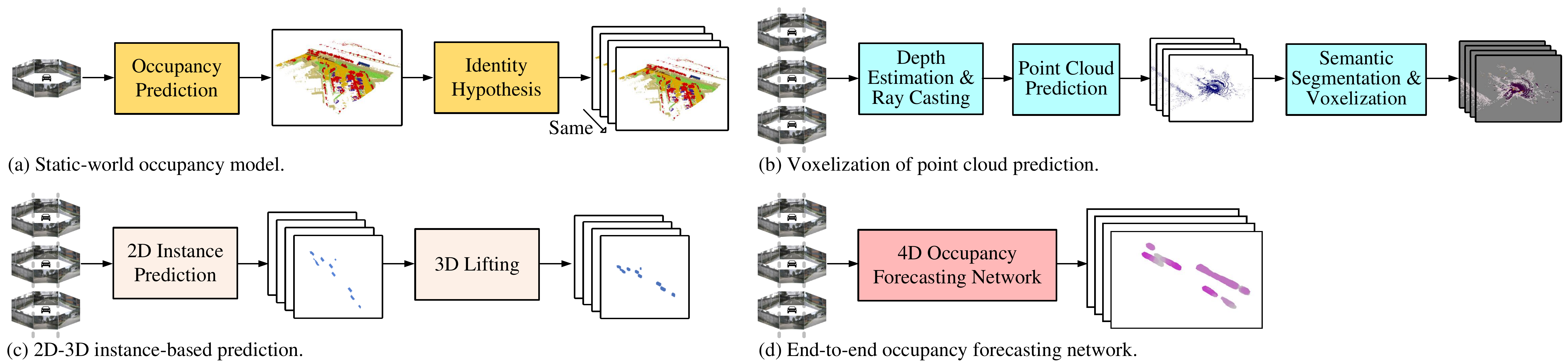}
  \vspace{-0.3cm}
  \caption{Four types of baselines are proposed in the \name{} benchmark from the extension of occupancy prediction, point cloud prediction, and 2D instance prediction, as well as our end-to-end 4D occupancy forecasting network.}
  \label{fig:baseline_methods}
  \vspace{-0.5cm}
\end{figure*}

\subsection{Baselines}
\label{sec:baselines}

We propose four methods as baselines in \name{} to assist future comparison for the camera-only 4D occupancy forecasting task as shown in~\figref{fig:baseline_methods}.

\textbf{Static-world occupancy model.} The existing camera-based occupancy prediction approaches \cite{wang2023openoccupancy,huang2023tri,tong2023scene,li2023voxformer,wei2023surroundocc,li2023stereovoxelnet} can only estimate the present occupancy grids based on the current observation. 
Therefore, one of the most straightforward baselines is to assume the environment remains static over a short time interval. Thus, we can use the present estimated occupancy grids as predictions for all future time steps based on the static-world hypothesis, as illustrated in~\figref{fig:baseline_methods}\,a. 

\textbf{Voxelization of point cloud prediction.} Another type of baseline can be the occupancy grid voxelization based on the point clouds forecasting results from existing point clouds prediction methods~\cite{fan2019pointrnn,lu2021monet,mersch2022self,luo2023pcpnet}. 
Here, we use surround-view depth estimation to generate depth maps across multiple cameras, followed by ray casting to generate 3D point clouds, which is applied with point cloud prediction to obtain predicted future pseudo points.
Based on that, we then apply point-based semantic segmentation~\cite{Zhu_2021_CVPR,chen2021mos,li2022semantic} to obtain movable and static labels for each voxel, resulting in the final occupancy predictions (see \figref{fig:baseline_methods}\,b). 

\textbf{2D-3D instance-based prediction.} Many off-the-shelf 2D BEV-based instance prediction methods \cite{hu2021fiery, ijcai2023p120, wu2020motionnet, mahjourian2022occupancy, hendy2020fishing} can forecast semantics for a near future with surround-view camera images. The third type of baseline is to obtain forecasted \gmo{} in 3D space by replicating the BEV occupancy grids along the z-axis to the height of the vehicle, as shown in~\figref{fig:baseline_methods}\,c. It can be seen that this baseline assumes that the driving surface is flat and all moving objects have the same height. We do not evaluate this baseline on forecasting \gso{} since boosting 2D results by replication is unsuitable for simulating large-scale backgrounds with much more complex structures compared to \gmo.

\textbf{End-to-end occupancy forecasting network.} None of the above baselines can directly predict the future occupancy state of 3D space. They all need additional post-processing based on certain hypotheses to extend and transform the existing results into 4D occupancy forecasting, inevitably introducing inherent artifacts. To fill this gap, we propose a novel approach shown in~\figref{fig:baseline_methods}\,d to achieve camera-only 4D occupancy forecasting in an end-to-end manner, introduced in detail in the next section. 

\section{End-to-End 4D Occupancy Forecasting}
\label{sec:our_network}

\begin{figure*}
  \centering
  \includegraphics[width=0.85\linewidth]{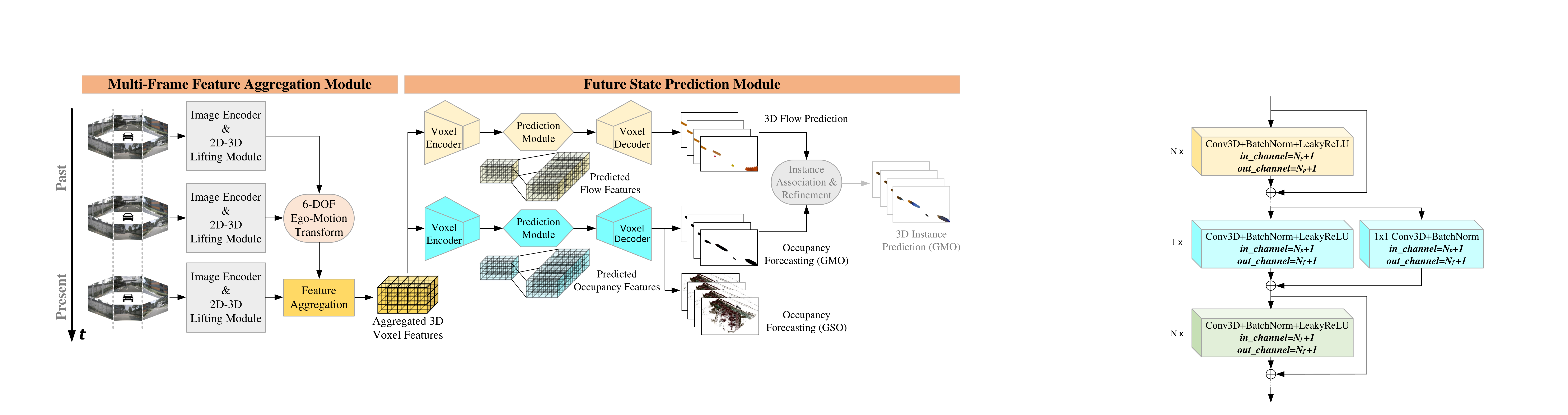}
  \vspace{-0.2cm}
  \caption{System overview of our proposed \netname.}
  \label{fig:our_model_all}
  \vspace{-0.5cm}
\end{figure*}

To our best knowledge, no existing camera-only 4D occupancy forecasting baseline is capable of simultaneously predicting future occupancy and extracting 3D general objects in an end-to-end fashion.
In this paper, we introduce a novel end-to-end spatio-temporal network dubbed \netname, depicted in~\figref{fig:our_model_all}. \netname{} receives sequential past surround-view camera images to predict the present and future occupancy states. It utilizes the multi-frame feature aggregation module to extract warped 3D voxel features and the future state prediction module to forecast future occupancy as well as 3D backward centripetal flow.

\subsection{Multi-Frame Feature Aggregation Module}
The multi-frame feature aggregation module takes a sequence of past surround-view camera images as input and employs an image encoder backbone to extract 2D features. These 2D features are subsequently lifted and integrated into 3D voxel features by the 2D-3D lifting module. All the resulting 3D feature volumes are transformed to the current coordinate system through the application of 6-DOF ego-car poses, yielding the aggregated feature $F_p \in \mathbb{R}^{(N_p+1)c \times h \times w \times l}$. Here, we collapse the time and feature dimensions into one dimension to implement the following 3D spatiotemporal convolution. Subsequently, we concatenate it with the 6-DOF relative ego-car poses between adjacent frames, leading to the motion-aware feature $F_{pm} \in \mathbb{R}^{(N_p+1)(c+6) \times h \times w \times l}$. 

\subsection{Future State Prediction Module}
With the motion-aware feature aggregated from sequential features as input, the future state prediction module uses two heads to forecast future occupancy as well as motion of the grids simultaneously. Firstly, a voxel encoder downsamples $F_{pm}$ to multi-scale features $F_{pm}^i  \in \mathbb{R}^{(N_p+1)c_i \times \frac{h}{2^{i}} \times \frac{w}{2^{i}} \times \frac{l}{2^{i}}}$, where $i=0, 1, 2, 3$. Then, the prediction module expands the channel dimension of each $F_{pm}^i$ to $(N_f+1)c_i$ using stacked 3D residual convolutional blocks (see \secref{sec:ocfnet_model_details} in supplementary materials), resulting in $F_{pf}^i  \in \mathbb{R}^{(N_f+1)c_i \times \frac{h}{2^{i}} \times \frac{w}{2^{i}} \times \frac{l}{2^{i}}}$. They are further concatenated with the feature upsampled by a voxel decoder, after which a softmax function is exploited in the occupancy forecasting head to produce the coarse occupancy feature $F_{f}^{occ} \in \mathbb{R}^{(N_f+1) \times cls \times h \times w \times l}$. In the flow prediction head, an additional $1\times 1$ convolutional layer instead of the softmax function is utilized to produce the coarse flow feature $F_{f}^{flow} \in \mathbb{R}^{(N_f+1) \times 3 \times h \times w \times l}$. Lastly, we utilize trilinear interpolation on $F_{f}^{occ}$ and $F_{f}^{flow}$, and an additional argmax function on the occupancy state dimension to generate the final occupancy estimation $\hat{\mathbf{O}}_{t} \in \mathbb{R}^{(N_f+1)  \times H \times W \times L}$ and flow-based motion prediction  $\hat{\mathbf{M}}_{t} \in \mathbb{R}^{(N_f+1) \times 3 \times H \times W \times L}$.
Here, we need to estimate the present and forecast the future occupancy with semantics of general objects simultaneously according to the evaluation protocol described in \secref{sec:eval_proto}. In addition, \netname{} not only forecasts occupancy but also predicts 3D backward centripetal flow as grid motion within the space, which can be utilized to achieve instance prediction (see \secref{sec:3d_instance_prediction} in supplementary materials).

\subsection{Loss function}
We use cross-entropy loss as the occupancy forecasting loss $L_{occ}$ and use smooth $l_1$ distance as the flow prediction loss $L_{flow}$. The explicit depth loss $L_{depth}$ \cite{li2023bevdepth} is also used as the previous work \cite{wang2023openoccupancy} suggests, but here it is only calculated for supervising the present occupancy ($t=0$) to improve training efficiency and decrease memory consumption. The overall loss for training \netname{} is given by
\begin{small}
\begin{align}
L_{all}=\frac{1}{N_f+1}\Big(\sum_{t=0}^{N_f}\lambda_1L_{occ}&(\hat{\mathbf{O}}_t, \mathbf{O}_t)+\lambda_2L_{flow}(\hat{\mathbf{M}}_t, \mathbf{M}_t)\Big)  \nonumber \\
	&+\lambda_3L_{depth}(\hat{\mathbf{D}}_0, \mathbf{D}_0),
\label{eq:loss_all}
\end{align}
\end{small}%
where $\hat{\mathbf{D}}_0, \mathbf{D}_0$ are the depth image estimated by the 2D-3D Lifting module and ground-truth range image projected from LiDAR data respectively. $\lambda_1$, $\lambda_2$, and $\lambda_3$ are the weights to balance the optimization for occupancy forecasting, flow prediction, and depth reconstruction.

\section{Experiments on \name}
\label{sec:exp}

Using the proposed \name{} benchmark, we evaluate the occupancy estimation and forecasting performance of the proposed baselines, including our \netname, for four tasks in autonomous driving scenarios. 

\begin{figure*}
  \centering
  \includegraphics[width=1.0\linewidth]
  {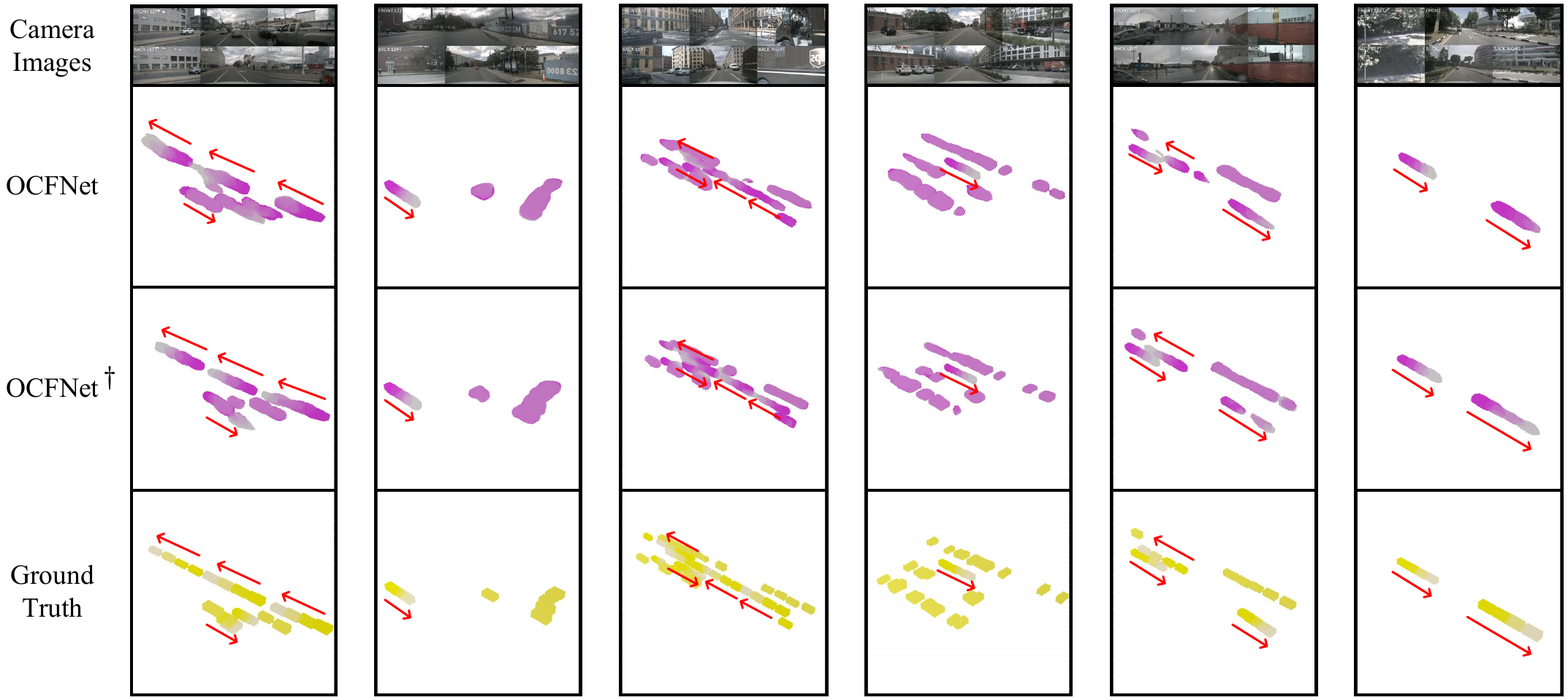}
  \vspace{-0.3cm}
  \caption{Visualization of forecasting inflated GMO by our proposed \netname{}. The prediction results and ground truth from timestamps 1 to $N_f$ are assigned colors from dark to light. The motion trend of each moving object is represented by red arrows.}
  \label{fig:viz_ocf}
  \vspace{-0.3cm}
\end{figure*}

\subsection{Experimental Setups}

\textbf{Dataset details.} Following~\cite{wang2023openoccupancy,ijcai2023p120}, we use 700 out of 850 scenes with ground-truth annotations in the nuScenes and nuScenes-Occupancy datasets, and 130 out of 180 scenes in the Lyft-Level5 for training the proposed baselines and our \netname. The remaining scenes are used for evaluation. The length $N$ of each sequence in our benchmark is set to 7 ($N_p=2$ and $N_f=4$), which means we use three observations, including the present one, to forecast occupancy in four incoming time steps.  Because nuScenes is annotated at 2\,Hz while Lyft-Level5 is annotated at 5\,Hz, we report the forecasting performance with different time intervals. The predefined range of each sequence is set as [-51.2\,m, 51.2\,m] for x-axis and y-axis, and [-5\,m, 3\,m] for z-axis. The voxel resolution is 0.2\,m, leading to occupancy grids with the size of $512\times 512\times 40$ in the present coordinate system of each sequence. After the data reorganizing of our \name{} benchmark, the number of sequences for training and test are 23930 and 5119 in nuScenes and nuScenes-Occupancy, and 15720 and 5880 in Lyft-Level5. 

\textbf{Baseline setups.} We choose the state-of-the-art camera-based approaches as the outset of each baseline proposed in \secref{sec:baselines}. For the static-world occupancy model, we use the camera baseline of OpenOccupancy~\cite{wang2023openoccupancy} (OpenOccupancy-C) to estimate the occupancy state of the present frame, which is then regarded as the prediction of all the future time steps. For the voxelization of point cloud prediction, we use SurroundDepth~\cite{wei2023surrounddepth} to estimate continuous surrounding depth maps, which are then downsampled to generate pseudo point clouds by ray casting. Based on sequential pseudo point clouds input, we then use PCPNet~\cite{luo2023pcpnet} to forecast incoming 3D point clouds, followed by Cylinder3D~\cite{Zhu_2021_CVPR} to extract point-level \gmo{} and \gso{} labels, and further voxelize the results into occupancy grids (SPC). For the 2D-3D instance-based prediction, we choose PowerBEV~\cite{ijcai2023p120} to forecast occupancy semantics on BEV and then lift the 2D results to 3D space (PowerBEV-3D). As to our proposed \netname, we directly implement 4D occupancy forecasting end-to-end. Notably, PowerBEV is trained by the 2D ground-truth semantics and 2D flow projected to the BEV plane. Besides, only PowerBEV and \netname{} are trained with flow annotations from \name{} simultaneously since they both have the flow head. To show that our proposed \netname{} can generate good forecasted results even seeing limited training data, we report the performance of \netname{} only trained on $\frac{1}{6}$ training sequences as well as the performance of the one trained on all training sequences (\netname$^{\dag}$).
OpenOccupancy-C, PowerBEV, and \netname{} are trained for 15 epochs using AdamW optimizer \cite{kingma2014adam} with an initial learning rate 3e-4 and a weight decay of 0.01. SurroundDepth and Cylinder3D used in the point cloud prediction baseline are fine-tuned as their open sources suggest. PCPNet is firstly pretrained by range loss for 40 epochs using the same optimizer, but the initial learning rate is set to 1e-3. After that, it is further fine-tuned by Chamfer distance loss \cite{fan2017point} for 10 epochs with a learning rate of 6e-4. All the networks mentioned above are trained with a batch size of 8 on 8 A100 GPUs. More details about the model parameters of our \netname{} are provided in supplementary \secref{sec:ocfnet_model_details}.

\subsection{4D Occupancy Forecasting Assessment}
\label{sec:assessment}


\textbf{Evaluation on forecasting inflated GMO.} Results of the first task, forecasting inflated \gmo{} on nuScenes and Lyft-Level5, are presented in~\tabref{tab:inflated_gmo}. Here, OpenOccupancy-C, PowerBEV, and \netname{} are trained only with inflated \gmo{} labels, while PCPNet is trained by holistic point clouds. As shown, \netname{} and \netname$^{\dag}$ outperform all other baselines, surpassing the BEV-based method by 12.4\% and 13.3\% in $\text{IoU}_f$ and $\tilde{\text{IoU}}_f$ on nuScenes. On Lyft-Level5, our \netname{} and \netname$^{\dag}$ consistently outperforms PowerBEV-3D by 20.8\% and 21.8\% in $\text{IoU}_f$ and $\tilde{\text{IoU}}_f$. In addition,
\figref{fig:viz_ocf} shows the results of nuScenes \gmo{} occupancy forecasted by our \netname{} and CFNet$^{\dag}$, which indicates that \netname{} trained only with limited data can still capture the motion of \gmo{} occupancy grids reasonably. The visualization on Lyft-Level5 is shown in supplementary \secref{sec:viz_on_lyft}. The baseline SPC cannot work well for the present frame and even tends to fail while forecasting future occupancy state. This is because movable objects are labeled as the inflated dense voxel grids in this task, while the voxelization of PCPNet outputs is from sparse point-level prediction. In addition, the shape of the predicted objects loses consistency significantly in future time steps. The performance of OpenOccupancy-C is much better than that of the point cloud prediction baseline but still has a weak ability to estimate present occupancy and forecast future occupancy compared to PowerEBV-3D and \netname. 

\begin{table}[t]
\scriptsize
\setlength{\tabcolsep}{2pt}
\center
\renewcommand\arraystretch{1.1}
\caption{Comparison of performance on forecasting inflated \gmo}
\vspace{-0.2cm}
\begin{tabular}{l|ccc|ccc}
\toprule
\multicolumn{1}{l|}{\multirow{2}{*}{approach}}   & \multicolumn{3}{c|}{nuScenes} & \multicolumn{3}{c}{Lyft-Level5} \\ \cmidrule{2-7} 
\multicolumn{1}{c|}{}                                                                               & $\text{IoU}_c$    & $\text{IoU}_f$ (2\,s) & $\tilde{\text{IoU}}_f$  & $\text{IoU}_c$    & $\text{IoU}_f$ (0.8\,s)  & $\tilde{\text{IoU}}_f$   \\ \cmidrule{1-7}
OpenOccupancy-C \cite{wang2023openoccupancy}                                                                    & 12.17      & 11.45      &  11.74    & 14.01       & 13.53     & 13.71          \\
SPC \cite{wei2023surrounddepth,luo2023pcpnet,Zhu_2021_CVPR}                                                           & 1.27        & failed      &  failed  & 1.42        & failed         & failed    \\ 
PowerBEV-3D \cite{ijcai2023p120}                                                                          & 23.08      & 21.25          & 21.86   & 26.19      & 24.47           & 25.06   \\ \cmidrule{1-7}
\netname{} (ours)                                                                          & 27.86           & 23.89     & 24.77     & 32.12      & 29.56      & 30.53        \\
\netname$^{\dag}$ (ours)                                                                                                     &  \textbf{31.30}          &  \textbf{26.82 }        & \textbf{27.98}     & \textbf{36.41}           & \textbf{33.56}        &  \textbf{34.60}          \\ \bottomrule
\multicolumn{7}{l}{SPC: SurroundDepth \cite{wei2023surrounddepth} + PCPNet \cite{luo2023pcpnet} + Cylinder3D \cite{Zhu_2021_CVPR}}
\end{tabular}
\label{tab:inflated_gmo}
\vspace{-0.6cm}
\end{table}

\textbf{Evaluation on forecasting fine-grained GMO.} We further report the occupancy estimation and forecasting performance on fine-grained general movable objects with nuScenes-Occupancy (the second-level task). 
In~\tabref{tab:fine_gmo}, we exhibit how the IoU of the forecasted objects changes once the \gmo{} annotations have fine-grained voxel format rather than the inflated one in the first-level task for training as well as evaluation. 
It can be seen that the IoU of \gmo{} forecasted by all the methods except the point cloud prediction baseline decreases significantly because it is rather difficult to predict sophisticated moving 3D structures using past continuous camera images. In contrast, SPC presents slightly better performance compared to the results in \tabref{tab:inflated_gmo} since the ground-truth labels are also fine-grained and sparser than the counterparts in the first-level task. However, due to the loss of shape consistency, it still has the worst performance among the baselines. Besides, we can also see in \tabref{tab:fine_gmo} that \netname{} and \netname$^{\dag}$ still have the best performance. This experiment reveals the reason why \name{} suggests the inflated labels for \gmo{} annotation in the occupancy forecasting task: Forecasting sophisticated future 3D structures of movable objects only using camera images is very difficult while forecasting inflated \gmo{} potentially promotes more reliable and safer navigation in autonomous driving applications.

\begin{table}[t] 
\scriptsize
\setlength{\tabcolsep}{12pt}
\center
\renewcommand\arraystretch{1.1}
\caption{Comparison on forecasting fine-grained \gmo{}}
\vspace{-0.2cm}
\begin{tabular}{l|ccc}
\toprule
\multirow{2}{*}{approach} & \multicolumn{3}{c}{nuScenes-Occupancy} \\ \cmidrule{2-4} 
                          & $\text{IoU}_{c}$           & $\text{IoU}_{f}$ (2\,s)     & $\tilde{\text{IoU}}_{f}$    \\ \cmidrule{1-4} 
OpenOccupancy-C \cite{wang2023openoccupancy}           & 10.82             & 8.02            & 8.53   \\ 
SPC \cite{wei2023surrounddepth,luo2023pcpnet,Zhu_2021_CVPR}        & 5.85              & 1.08       &  1.12          \\ 
PowerBEV-3D \cite{ijcai2023p120}                  & 5.91              & 5.25         & 5.49      \\ \cmidrule{1-4} 
\netname{} (ours)             & 10.15      & 8.35          & 8.69      \\
\netname$^{\dag}$ (ours)            & \textbf{11.45}                  & \textbf{9.68}               &  \textbf{10.10}    \\ \bottomrule
\end{tabular}
\label{tab:fine_gmo}
\end{table}
\begin{table}[t] 
\scriptsize
\setlength{\tabcolsep}{3pt}
\center
\renewcommand\arraystretch{1.1}
\caption{Comparison of performance on forecasting inflated \gmo, fine-grained GSO, and free space simultaneously}

\begin{tabular}{l|ccccccc}
\toprule
\multirow{2}{*}{approach} & \multicolumn{3}{c}{$\text{IoU}_c$}                                        & \multicolumn{3}{c}{$\text{IoU}_f$ (2\,s)}                                          & \multicolumn{1}{c}{$\tilde{\text{IoU}}_f$}                                         \\ \cmidrule{2-8} 
                          & \rotatebox{60}{GMO}                  & \rotatebox{60}{GSO}                  & \rotatebox{60}{mean}                 & \rotatebox{60}{GMO}                  & \rotatebox{60}{GSO}                  & \rotatebox{60}{mean}              & \rotatebox{60}{GMO}   \\ \cmidrule{1-8}
OpenOccupancy-C\,\cite{wang2023openoccupancy}           & 13.53                & 16.86                & 15.20                & 12.67                 & 17.09                     & 14.88                &  12.97      \\ 
SPC \cite{wei2023surrounddepth,luo2023pcpnet,Zhu_2021_CVPR}      & 1.27                 & 3.29                 & 2.28                 & failed                 & 1.40                 & --               & failed   \\ 
PowerBEV-3D \cite{ijcai2023p120}      & 23.08                 & --                 & --                 & 21.25                 & --                 & --               & 21.86   \\ \cmidrule{1-8}
\netname{} (ours)             & 26.41                & 16.95             & 21.68                     & 22.21                     & 17.14                     & 19.68                 & 23.06     \\ 
\netname$^{\dag}$ (ours)            & \textbf{29.84} & \textbf{17.72} & \textbf{23.78} & \textbf{25.53} & \textbf{17.81} & \textbf{21.67}     & \textbf{26.53} \\ \bottomrule
\end{tabular}
\label{tab:all_class}
\vspace{-0.5cm}
\end{table}


\textbf{Evaluation on forecasting inflated GMO, fine-grained GSO, and free space.} Next, we compare the performance of different methods on forecasting inflated general movable objects, fine-grained general static objects, and free space (the third-level task). Here, we do not report the GSO results from the 2D-3D instance-based prediction since the fine-grained 3D structure of static foreground and background objects cannot be approximately estimated by lifting 2D voxel grids to 3D space. The experimental results are shown in~\tabref{tab:all_class}. SPC remains the worst in this experiment where the IoU of inflated GMO is consistent with the results of~\tabref{tab:inflated_gmo}.
\netname{} and \netname$^{\dag}$ outperform OpenOccupancy-C significantly in terms of estimating GMO occupancy in both present moment and future time steps. It also can be seen that by aggregating features of multiple past frames, \netname$^{\dag}$ enhances the performance of GSO occupancy estimation of single-frame-based OpenOccupancy-C by 5.1\% and 4.2\% on IoU$_c$ and IoU$_f$ respectively. For OpenOccupancy-C and our \netname, the IoU values of future GSO slightly increase due to the jitter of ground truth annotations from nuScenes-Occupancy.  


\textbf{Evaluation on forecasting fine-grained GMO, fine-grained GSO, and free space.} In the fourth-level task, only OpenOccupancy-C and our \netname{} need to be retrained. As seen in \tabref{tab:all_class_fine}, OCFNet$^{\dag}$ remains the best performance against all the other approaches on forecasting fine-grained objects of interest. Compared to the results in \tabref{tab:fine_gmo}, the GMO forecasting performance of OpenOccupancy-C and our \netname{} drops slightly due to additional artifacts introduced by the fine-grained GSO class.

\begin{table}[t] 
\scriptsize
\setlength{\tabcolsep}{3pt}
\center
\renewcommand\arraystretch{1.1}
\caption{Comparison of performance on forecasting fine-grained \gmo, fine-grained GSO, and free space simultaneously}
\vspace{-0.2cm}
\begin{tabular}{l|ccccccc}
\toprule
\multirow{2}{*}{approach} & \multicolumn{3}{c}{$\text{IoU}_c$}                                        & \multicolumn{3}{c}{$\text{IoU}_f$ (2\,s)}                                          & \multicolumn{1}{c}{$\tilde{\text{IoU}}_f$}                                         \\ \cmidrule{2-8} 
                          & \rotatebox{60}{GMO}                  & \rotatebox{60}{GSO}                  & \rotatebox{60}{mean}                 & \rotatebox{60}{GMO}                  & \rotatebox{60}{GSO}                  & \rotatebox{60}{mean}              & \rotatebox{60}{GMO}   \\ \cmidrule{1-8}
OpenOccupancy-C\,\cite{wang2023openoccupancy}           & 9.62                &  17.21               &  13.42               & 7.41                 & 17.30                     & 12.36                & 7.86       \\ 
SPC \cite{wei2023surrounddepth,luo2023pcpnet,Zhu_2021_CVPR}      & 5.85                 & 3.29                & 4.57                 & 1.08                 & 1.40                 & 1.24               & 1.12   \\ 
PowerBEV-3D \cite{ijcai2023p120}      & 5.91                 & --                 & --                 & 5.25                 & --                 & --               & 5.49   \\ \cmidrule{1-8}
\netname{} (ours)             & 9.54                & 17.30             & 13.42                     &  8.23                    & 17.32                     & 12.78                 & 8.46     \\ 
\netname$^{\dag}$ (ours)            & \textbf{11.02}  & \textbf{17.79} & \textbf{14.41} & \textbf{9.20}  & \textbf{17.83}  & \textbf{13.52}    & \textbf{9.66}  \\ \bottomrule
\end{tabular}
\label{tab:all_class_fine}
\end{table}

\subsection{Ablation Study on Multi-Task Learning}
\label{sec:ab_on_flow}

In this experiment, we conduct an ablation study on the flow prediction head to present the enhancement from the multi-task learning scheme. 
As \tabref{tab:ab_flow} shows, the complete \netname{} enhances the one without the flow prediction head by around 4\% in both present and future occupancy estimation. 
The reason could be that 3D flow guides learning \gmo{} motion in each time interval, as shown in \secref{sec:3d_flow_prediction} in supplementary materials, and thus helps the model determine the change of occupancy estimation in the next timestamp. With this analysis, using 3D backward centripetal flow in our \name{} is suggested for future end-to-end 4D Occupancy forecasting models to achieve better forecasting performance.

\begin{table}[t] \scriptsize
\setlength{\tabcolsep}{6pt}
\center
\renewcommand\arraystretch{1.1}
\caption{Ablation study on flow prediction head}
\vspace{-0.2cm}
\begin{tabular}{l|cccccc}
\toprule
\multirow{2}{*}{approach} & $\text{IoU}_c$ & \multicolumn{4}{c}{$\text{IoU}_f$}        & $\tilde{\text{IoU}}_f$ \\ \cmidrule{2-7} 
                          &         & 0.5\,s & 1.0\,s & 1.5\,s & 2.0\,s &           \\ \cmidrule{1-7} 
\netname{} w/o flow           & 26.84        & 25.01       & 24.04       & 23.38       & 22.99       & 23.86         \\ 
\netname                    & \textbf{27.86}       & \textbf{25.95}       & \textbf{24.92}       & \textbf{24.33}       & \textbf{23.89}       & \textbf{24.77}           \\ \cmidrule{1-7}
improvement $\uparrow$                    & 3.8\%       & 3.8\%      & 3.7\%       & 4.1\%       & 3.9\%       & 3.8\%           \\ \bottomrule
\end{tabular}
\label{tab:ab_flow}
\vspace{-0.4cm}
\end{table}

\section{Conclusion}
\label{sec:conclusion}

In this paper, we propose a novel benchmark namely \name{} for the new task, camera-only 4D occupancy forecasting in autonomous driving applications. Specifically, we first establish the devised dataset in new format based on several publicly available datasets. Then the standardized evaluation protocol as well as four types of baselines are further proposed to provide basic reference in our \name{} benchmark. Moreover, we propose the first camera-based 4D occupancy forecasting network OCFNet to estimate future occupancy states in an end-to-end manner. Multiple experiments with four different tasks are conducted based on our \name{} benchmark to thoroughly evaluate the proposed baselines as well as our OCFNet. The experimental results show that OCFNet outperforms all the baselines and can still produce reasonable future occupancy even seeing limited training data. 

\textbf{Insights:} By comparing four different types of baselines, we demonstrated that end-to-end spatiotemporal network could be the most promising research direction for camera-only occupancy forecasting. Besides, using inflated GMO annotation and additional 3D backward centripetal flow is also verified to be beneficial for 4D occupancy forecasting. 

\textbf{Limitation and future work:} While notable results have been achieved by our OCFNet, camera-only 4D occupancy forecasting remains challenging, especially for predicting over longer time intervals with many moving objects. Our \name{} benchmark and comprehensive analysis aim to enhance understanding of the strengths and limitations of current occupancy perception models.
We envision this benchmark as a valuable tool for evaluation, and our OCFNet can serve as a foundational codebase for future research in the task of 4D occupancy forecasting.


%



\ifCLASSOPTIONcaptionsoff
  \newpage
\fi



%


\bibliographystyle{unsrt}
\bibliography{main}

\begin{thebibliography}{10}

\bibitem{liu2022petr}
Yingfei Liu, Tiancai Wang, Xiangyu Zhang, and Jian Sun.
\newblock Petr: Position embedding transformation for multi-view 3d object detection.
\newblock In {\em ECCV}, pages 531--548, 2022.

\bibitem{reading2021categorical}
Cody Reading, Ali Harakeh, Julia Chae, and Steven~L Waslander.
\newblock Categorical depth distribution network for monocular 3d object detection.
\newblock In {\em CVPR}, pages 8555--8564, 2021.

\bibitem{wang2023yolov7}
Chien-Yao Wang, Alexey Bochkovskiy, and Hong-Yuan~Mark Liao.
\newblock Yolov7: Trainable bag-of-freebies sets new state-of-the-art for real-time object detectors.
\newblock In {\em CVPR}, pages 7464--7475, 2023.

\bibitem{zhou2022mogde}
Yunsong Zhou, Quan Liu, Hongzi Zhu, Yunzhe Li, Shan Chang, and Minyi Guo.
\newblock Mogde: Boosting mobile monocular 3d object detection with ground depth estimation.
\newblock {\em NeurIPS}, 35:2033--2045, 2022.

\bibitem{xie2021segformer}
Enze Xie, Wenhai Wang, Zhiding Yu, Anima Anandkumar, Jose~M Alvarez, and Ping Luo.
\newblock Segformer: Simple and efficient design for semantic segmentation with transformers.
\newblock {\em NeurIPS}, 34:12077--12090, 2021.

\bibitem{Li_2022_CVPR}
Liulei Li, Tianfei Zhou, Wenguan Wang, Jianwu Li, and Yi~Yang.
\newblock Deep hierarchical semantic segmentation.
\newblock In {\em CVPR}, pages 1246--1257, June 2022.

\bibitem{long2015fully}
Jonathan Long, Evan Shelhamer, and Trevor Darrell.
\newblock Fully convolutional networks for semantic segmentation.
\newblock In {\em CVPR}, pages 3431--3440, 2015.

\bibitem{chen2017deeplab}
Liang-Chieh Chen, George Papandreou, Iasonas Kokkinos, Kevin Murphy, and Alan~L Yuille.
\newblock Deeplab: Semantic image segmentation with deep convolutional nets, atrous convolution, and fully connected crfs.
\newblock {\em IEEE TPAMI}, 40(4):834--848, 2017.

\bibitem{voedisch23codeps}
Niclas Vödisch, Kürsat Petek, Wolfram Burgard, and Abhinav Valada.
\newblock Codeps: Online continual learning for depth estimation and panoptic segmentation.
\newblock {\em RSS}, 2023.

\bibitem{hu2023you}
Jie Hu, Linyan Huang, Tianhe Ren, Shengchuan Zhang, Rongrong Ji, and Liujuan Cao.
\newblock You only segment once: Towards real-time panoptic segmentation.
\newblock In {\em CVPR}, pages 17819--17829, 2023.

\bibitem{li2023point2mask}
Wentong Li, Yuqian Yuan, Song Wang, Jianke Zhu, Jianshu Li, Jian Liu, and Lei Zhang.
\newblock Point2mask: Point-supervised panoptic segmentation via optimal transport.
\newblock In {\em ICCV}, pages 572--581, 2023.

\bibitem{cheng2020panoptic}
Bowen Cheng, Maxwell~D Collins, Yukun Zhu, Ting Liu, Thomas~S Huang, Hartwig Adam, and Liang-Chieh Chen.
\newblock Panoptic-deeplab: A simple, strong, and fast baseline for bottom-up panoptic segmentation.
\newblock In {\em CVPR}, pages 12475--12485, 2020.

\bibitem{wang2023openoccupancy}
Xiaofeng Wang, Zheng Zhu, Wenbo Xu, Yunpeng Zhang, Yi~Wei, Xu~Chi, Yun Ye, Dalong Du, Jiwen Lu, and Xingang Wang.
\newblock Openoccupancy: A large scale benchmark for surrounding semantic occupancy perception.
\newblock In {\em ICCV}, pages 17850--17859, October 2023.

\bibitem{huang2023tri}
Yuanhui Huang, Wenzhao Zheng, Yunpeng Zhang, Jie Zhou, and Jiwen Lu.
\newblock Tri-perspective view for vision-based 3d semantic occupancy prediction.
\newblock In {\em CVPR}, pages 9223--9232, 2023.

\bibitem{tong2023scene}
Wenwen Tong, Chonghao Sima, Tai Wang, Li~Chen, Silei Wu, Hanming Deng, Yi~Gu, Lewei Lu, Ping Luo, Dahua Lin, et~al.
\newblock Scene as occupancy.
\newblock In {\em ICCV}, pages 8406--8415, 2023.

\bibitem{li2023voxformer}
Yiming Li, Zhiding Yu, Christopher Choy, Chaowei Xiao, Jose~M Alvarez, Sanja Fidler, Chen Feng, and Anima Anandkumar.
\newblock Voxformer: Sparse voxel transformer for camera-based 3d semantic scene completion.
\newblock In {\em CVPR}, pages 9087--9098, 2023.

\bibitem{wei2023surroundocc}
Yi~Wei, Linqing Zhao, Wenzhao Zheng, Zheng Zhu, Jie Zhou, and Jiwen Lu.
\newblock Surroundocc: Multi-camera 3d occupancy prediction for autonomous driving.
\newblock In {\em ICCV}, pages 21729--21740, 2023.

\bibitem{ding2021epsilon}
Wenchao Ding, Lu~Zhang, Jing Chen, and Shaojie Shen.
\newblock Epsilon: An efficient planning system for automated vehicles in highly interactive environments.
\newblock {\em TRO}, 38(2):1118--1138, 2021.

\bibitem{ding2019safe}
Wenchao Ding, Lu~Zhang, Jing Chen, and Shaojie Shen.
\newblock Safe trajectory generation for complex urban environments using spatio-temporal semantic corridor.
\newblock {\em RA-L}, 4(3):2997--3004, 2019.

\bibitem{song2020pip}
Haoran Song, Wenchao Ding, Yuxuan Chen, Shaojie Shen, Michael~Yu Wang, and Qifeng Chen.
\newblock Pip: Planning-informed trajectory prediction for autonomous driving.
\newblock In {\em ECCV}, pages 598--614, 2020.

\bibitem{hu2021fiery}
Anthony Hu, Zak Murez, Nikhil Mohan, Sof{\'\i}a Dudas, Jeffrey Hawke, Vijay Badrinarayanan, Roberto Cipolla, and Alex Kendall.
\newblock Fiery: Future instance prediction in bird's-eye view from surround monocular cameras.
\newblock In {\em ICCV}, pages 15273--15282, 2021.

\bibitem{ijcai2023p120}
Peizheng Li, Shuxiao Ding, Xieyuanli Chen, Niklas Hanselmann, Marius Cordts, and Juergen Gall.
\newblock Powerbev: A powerful yet lightweight framework for instance prediction in bird’s-eye view.
\newblock In {\em IJCAI}, pages 1080--1088, 8 2023.

\bibitem{wu2020motionnet}
Pengxiang Wu, Siheng Chen, and Dimitris~N Metaxas.
\newblock Motionnet: Joint perception and motion prediction for autonomous driving based on bird's eye view maps.
\newblock In {\em CVPR}, pages 11385--11395, 2020.

\bibitem{mahjourian2022occupancy}
Reza Mahjourian, Jinkyu Kim, Yuning Chai, Mingxing Tan, Ben Sapp, and Dragomir Anguelov.
\newblock Occupancy flow fields for motion forecasting in autonomous driving.
\newblock {\em RA-L}, 7(2):5639--5646, 2022.

\bibitem{hendy2020fishing}
Noureldin Hendy, Cooper Sloan, Feng Tian, Pengfei Duan, Nick Charchut, Yuesong Xie, Chuang Wang, and James Philbin.
\newblock Fishing net: Future inference of semantic heatmaps in grids.
\newblock In {\em CVPRW}, 2020.

\bibitem{khurana2023point}
Tarasha Khurana, Peiyun Hu, David Held, and Deva Ramanan.
\newblock Point cloud forecasting as a proxy for 4d occupancy forecasting.
\newblock In {\em CVPR}, pages 1116--1124, 2023.

\bibitem{khurana2022differentiable}
Tarasha Khurana, Peiyun Hu, Achal Dave, Jason Ziglar, David Held, and Deva Ramanan.
\newblock Differentiable raycasting for self-supervised occupancy forecasting.
\newblock In {\em ECCV}, pages 353--369, 2022.

\bibitem{toyungyernsub2022dynamics}
Maneekwan Toyungyernsub, Esen Yel, Jiachen Li, and Mykel~J Kochenderfer.
\newblock Dynamics-aware spatiotemporal occupancy prediction in urban environments.
\newblock In {\em IROS}, pages 10836--10841, 2022.

\bibitem{caesar2020nuscenes}
Holger Caesar, Varun Bankiti, Alex~H Lang, Sourabh Vora, Venice~Erin Liong, Qiang Xu, Anush Krishnan, Yu~Pan, Giancarlo Baldan, and Oscar Beijbom.
\newblock nuscenes: A multimodal dataset for autonomous driving.
\newblock In {\em CVPR}, pages 11621--11631, 2020.

\bibitem{lyft2019}
R.~Kesten, M.~Usman, J.~Houston, T.~Pandya, K.~Nadhamuni, A.~Ferreira, M.~Yuan, B.~Low, A.~Jain, P.~Ondruska, S.~Omari, S.~Shah, A.~Kulkarni, A.~Kazakova, C.~Tao, L.~Platinsky, W.~Jiang, and V.~Shet.
\newblock Lyft level 5 perception dataset 2020, 2019.

\bibitem{cao2022monoscene}
Anh-Quan Cao and Raoul de~Charette.
\newblock Monoscene: Monocular 3d semantic scene completion.
\newblock In {\em CVPR}, pages 3991--4001, 2022.

\bibitem{pan2023uniocc}
Mingjie Pan, Li~Liu, Jiaming Liu, Peixiang Huang, Longlong Wang, Shanghang Zhang, Shaoqing Xu, Zhiyi Lai, and Kuiyuan Yang.
\newblock Uniocc: Unifying vision-centric 3d occupancy prediction with geometric and semantic rendering.
\newblock {\em arXiv preprint arXiv:2306.09117}, 2023.

\bibitem{tian2023occ3d}
Xiaoyu Tian, Tao Jiang, Longfei Yun, Yue Wang, Yilun Wang, and Hang Zhao.
\newblock Occ3d: A large-scale 3d occupancy prediction benchmark for autonomous driving.
\newblock {\em arXiv preprint arXiv:2304.14365}, 2023.

\bibitem{fan2019pointrnn}
Hehe Fan and Yi~Yang.
\newblock Pointrnn: Point recurrent neural network for moving point cloud processing.
\newblock {\em arXiv preprint arXiv:1910.08287}, 2019.

\bibitem{lu2021monet}
Fan Lu, Guang Chen, Zhijun Li, Lijun Zhang, Yinlong Liu, Sanqing Qu, and Alois Knoll.
\newblock Monet: Motion-based point cloud prediction network.
\newblock {\em TITS}, 23(8):13794--13804, 2021.

\bibitem{mersch2022self}
Benedikt Mersch, Xieyuanli Chen, Jens Behley, and Cyrill Stachniss.
\newblock Self-supervised point cloud prediction using 3d spatio-temporal convolutional networks.
\newblock In {\em CoRL}, pages 1444--1454, 2022.

\bibitem{luo2023pcpnet}
Zhen Luo, Junyi Ma, Zijie Zhou, and Guangming Xiong.
\newblock Pcpnet: An efficient and semantic-enhanced transformer network for point cloud prediction.
\newblock {\em RA-L}, 2023.

\bibitem{akan2022stretchbev}
Adil~Kaan Akan and Fatma G{\"u}ney.
\newblock Stretchbev: Stretching future instance prediction spatially and temporally.
\newblock In {\em ECCV}, pages 444--460, 2022.

\bibitem{zhang2022beverse}
Yunpeng Zhang, Zheng Zhu, Wenzhao Zheng, Junjie Huang, Guan Huang, Jie Zhou, and Jiwen Lu.
\newblock Beverse: Unified perception and prediction in birds-eye-view for vision-centric autonomous driving.
\newblock {\em arXiv preprint arXiv:2205.09743}, 2022.

\bibitem{chen20203d}
Xiaokang Chen, Kwan-Yee Lin, Chen Qian, Gang Zeng, and Hongsheng Li.
\newblock 3d sketch-aware semantic scene completion via semi-supervised structure prior.
\newblock In {\em CVPR}, pages 4193--4202, 2020.

\bibitem{li2020anisotropic}
Jie Li, Kai Han, Peng Wang, Yu~Liu, and Xia Yuan.
\newblock Anisotropic convolutional networks for 3d semantic scene completion.
\newblock In {\em CVPR}, pages 3351--3359, 2020.

\bibitem{roldao2020lmscnet}
Luis Roldao, Raoul de~Charette, and Anne Verroust-Blondet.
\newblock Lmscnet: Lightweight multiscale 3d semantic completion.
\newblock In {\em 3DV}, pages 111--119, 2020.

\bibitem{casas2021mp3}
Sergio Casas, Abbas Sadat, and Raquel Urtasun.
\newblock Mp3: A unified model to map, perceive, predict and plan.
\newblock In {\em CVPR}, pages 14403--14412, 2021.

\bibitem{chen2022auto}
Xieyuanli Chen, Benedikt Mersch, Lucas Nunes, Rodrigo Marcuzzi, Ignacio Vizzo, Jens Behley, and Cyrill Stachniss.
\newblock {Automatic Labeling to Generate Training Data for Online LiDAR-Based Moving Object Segmentation}.
\newblock {\em RA-L}, 7(3):6107--6114, 2022.

\bibitem{li2023stereovoxelnet}
Hongyu Li, Zhengang Li, Ne{\c{s}}et~{\"U}nver Akmandor, Huaizu Jiang, Yanzhi Wang, and Ta{\c{s}}k{\i}n Pad{\i}r.
\newblock Stereovoxelnet: Real-time obstacle detection based on occupancy voxels from a stereo camera using deep neural networks.
\newblock In {\em ICRA}, pages 4826--4833, 2023.

\bibitem{Zhu_2021_CVPR}
Xinge Zhu, Hui Zhou, Tai Wang, Fangzhou Hong, Yuexin Ma, Wei Li, Hongsheng Li, and Dahua Lin.
\newblock Cylindrical and asymmetrical 3d convolution networks for lidar segmentation.
\newblock In {\em CVPR}, pages 9939--9948, June 2021.

\bibitem{chen2021mos}
Xieyuanli Chen, Shijie Li, Benedikt Mersch, Louis Wiesmann, Jürgen Gall, Jens Behley, and Cyrill Stachniss.
\newblock Moving object segmentation in 3d lidar data: A learning-based approach exploiting sequential data.
\newblock {\em RA-L}, 6(4):6529--6536, 2021.

\bibitem{li2022semantic}
Shijie Li, Xieyuanli Chen, Yun Liu, Dengxin Dai, Cyrill Stachniss, and Juergen Gall.
\newblock Multi-scale interaction for real-time lidar data segmentation on an embedded platform.
\newblock {\em RA-L}, 7(2):738--745, 2022.

\bibitem{li2023bevdepth}
Yinhao Li, Zheng Ge, Guanyi Yu, Jinrong Yang, Zengran Wang, Yukang Shi, Jianjian Sun, and Zeming Li.
\newblock Bevdepth: Acquisition of reliable depth for multi-view 3d object detection.
\newblock In {\em AAAI}, volume~37, pages 1477--1485, 2023.

\bibitem{wei2023surrounddepth}
Yi~Wei, Linqing Zhao, Wenzhao Zheng, Zheng Zhu, Yongming Rao, Guan Huang, Jiwen Lu, and Jie Zhou.
\newblock Surrounddepth: Entangling surrounding views for self-supervised multi-camera depth estimation.
\newblock In {\em CoRL}, pages 539--549, 2023.

\bibitem{kingma2014adam}
Diederik~P Kingma and Jimmy Ba.
\newblock Adam: A method for stochastic optimization.
\newblock In {\em ICLR}, 2015.

\bibitem{fan2017point}
Haoqiang Fan, Hao Su, and Leonidas~J Guibas.
\newblock A point set generation network for 3d object reconstruction from a single image.
\newblock In {\em CVPR}, pages 605--613, 2017.

\bibitem{he2016deep}
Kaiming He, Xiangyu Zhang, Shaoqing Ren, and Jian Sun.
\newblock Deep residual learning for image recognition.
\newblock In {\em CVPR}, pages 770--778, 2016.

\bibitem{deng2009imagenet}
Jia Deng, Wei Dong, Richard Socher, Li-Jia Li, Kai Li, and Li~Fei-Fei.
\newblock Imagenet: A large-scale hierarchical image database.
\newblock In {\em CVPR}, pages 248--255. Ieee, 2009.

\bibitem{lin2017feature}
Tsung-Yi Lin, Piotr Doll{\'a}r, Ross Girshick, Kaiming He, Bharath Hariharan, and Serge Belongie.
\newblock Feature pyramid networks for object detection.
\newblock In {\em CVPR}, pages 2117--2125, 2017.

\bibitem{philion2020lift}
Jonah Philion and Sanja Fidler.
\newblock Lift, splat, shoot: Encoding images from arbitrary camera rigs by implicitly unprojecting to 3d.
\newblock In {\em ECCV}, pages 194--210, 2020.

\bibitem{kim2020video}
Dahun Kim, Sanghyun Woo, Joon-Young Lee, and In~So Kweon.
\newblock Video panoptic segmentation.
\newblock In {\em CVPR}, pages 9859--9868, 2020.

\end{thebibliography}

\clearpage
\begin{figure*}[t]
    \centering
    \LARGE
    \name: Benchmark for Camera-Only 4D Occupancy Forecasting \\ in Autonomous Driving Applications Supplementary Material
\end{figure*}

\renewcommand\thesection{\Alph{section}}
\setcounter{section}{0}

\section{Dataset Setup Details}
\label{sec:dataset_setup_details}

\begin{figure}[b]
  \centering
  \includegraphics[width=0.85\linewidth]{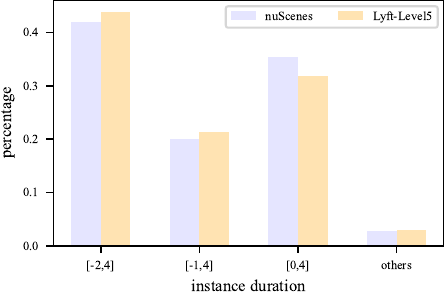}
  \caption{Instance duration on nuScenes and Lyft-Level5.}
  \label{fig:instance_duration}
\end{figure}

We provide more details about our new dataset format for our \name{} benchmark by presenting statistics on the instance duration $[t_{in}, t_{out}]$ after splitting the original nuScenes and Lyft-Level5 datasets to separate sequences mentioned in \secref{sec:dataset_in_new_format}. As shown in~\figref{fig:instance_duration}, most general movable objects (\gmo) appear in at least two historical observations and all future observations ($[-2,4]$ and $[-1,4]$) in our benchmark. The long instance duration leads to an effective training strategy for the occupancy forecasting model. Besides, over 30\% instances in the two datasets first appear in the current frame ($t=0$), which makes the model learn to forecast the object motion only according to their current location and surrounding conditions.

\begin{figure}[h]
  \centering
  \vspace{0.5cm}
  \includegraphics[width=0.96\linewidth]{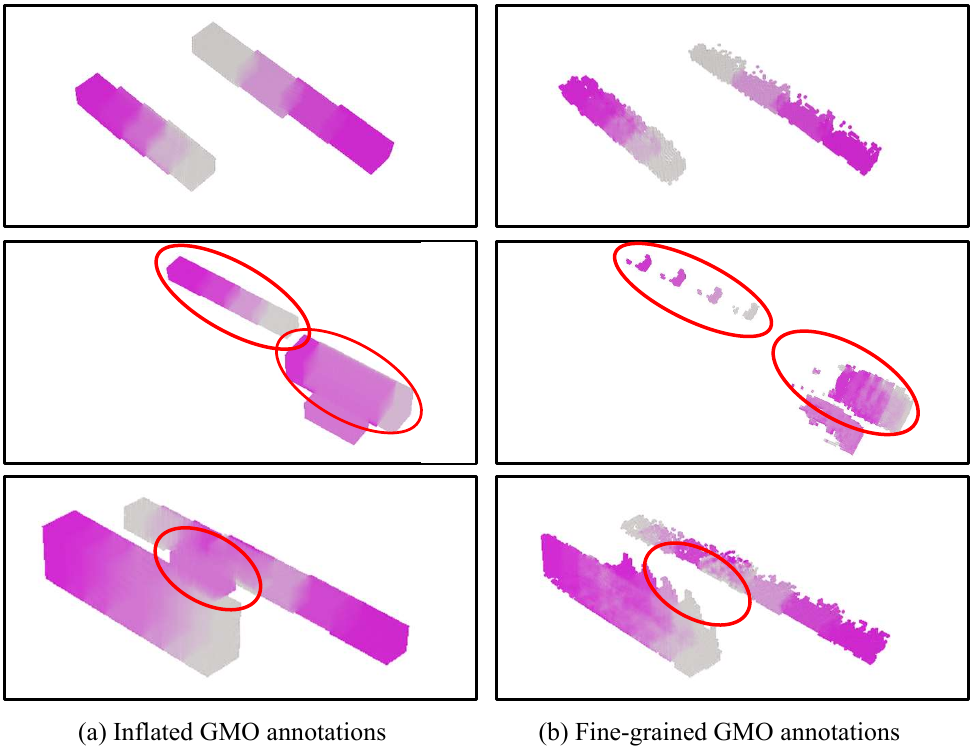}
  \caption{Comparison of GMO categories defined in Cam4DOcc.}
  \label{fig:show_inf_fine}
  \vspace{-0.5cm}
\end{figure}

In addition, we further provide a detailed illustration of inflated GMO and fine-grained GMO defined in our Cam4DOcc introduced in \secref{sec:eval_proto}, as shown in~\figref{fig:show_inf_fine}. Compared to the fine-grained labels, the inflated bounding-box-wise annotation overall provides more comprehensive training signals for the occupancy forecasting model. In addition, the motion of GMO with a structured format from the instance bounding box is easier to capture (validated in \secref{sec:assessment}). From the second row of \figref{fig:show_inf_fine} we can also see that sometimes fine-grained voxel annotation cannot accurately represent the sophisticated shape of GMO while the bounding-box-wise annotation can totally encompass the holistic GMO instance grids. The third row of \figref{fig:show_inf_fine} also presents that fine-grained annotation may miss some occluded objects compared to the original instance bounding box labels, affecting the rationality of the training and evaluation on these scenarios.
Therefore, Cam4DOcc suggests using inflated GMO annotations to train current-stage camera-based models for more reliable 4D occupancy forecasting and safer navigation in autonomous driving. We also hope that the preset tasks with fine-grained GMO labels in Cam4DOcc can be the foundation for developing more advanced camera-only 4D occupancy forecasting approaches in future research.

\section{OCFNet Model Details}
\label{sec:ocfnet_model_details}

\begin{figure}
  \centering
\includegraphics[width=0.9\linewidth]{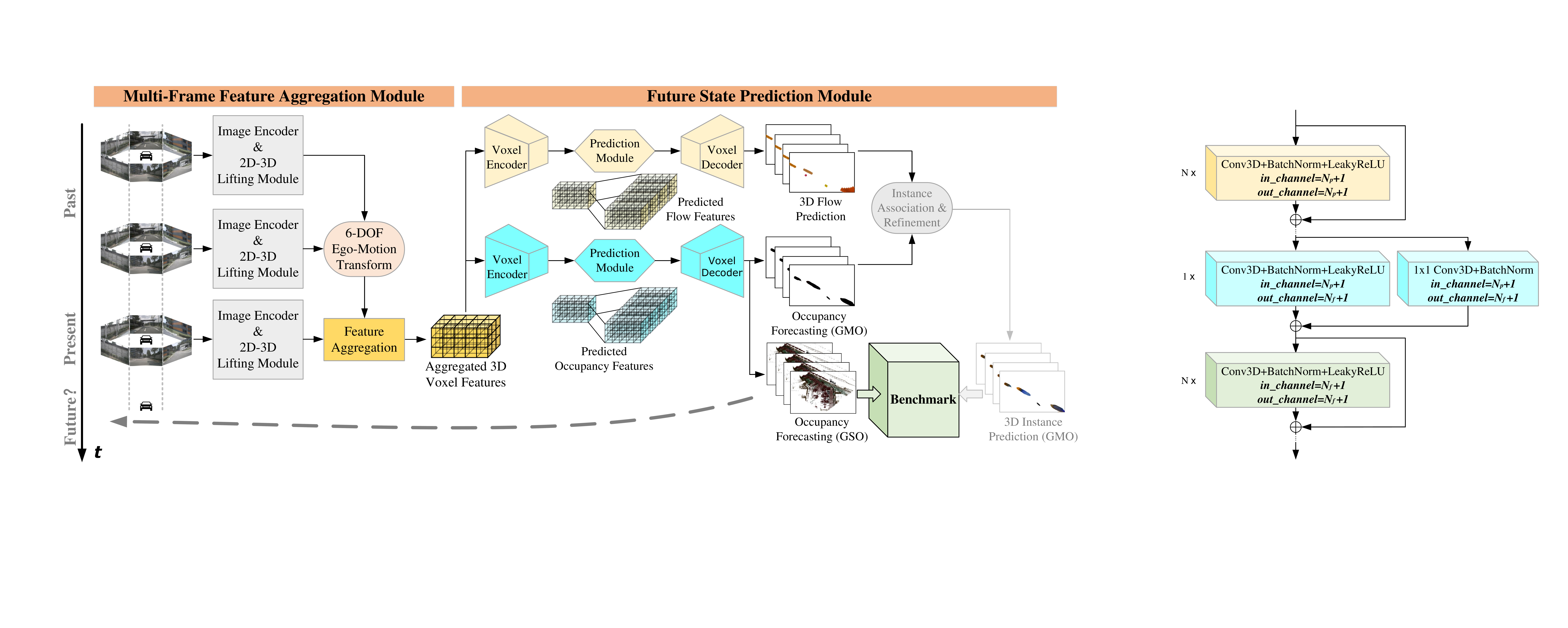}
  \caption{The prediction module in our OCFNet.}
  \label{fig:our_model_pred}
\end{figure}

Our proposed OCFNet receives 6 images with the size of $900\times 1600$ captured by surround-view cameras mounted on the vehicle. We use ResNet50~\cite{he2016deep} pretrained on ImageNet~\cite{deng2009imagenet} with FPN~\cite{lin2017feature} as the image encoder in OCFNet. LSS-based 2D-3D Lifting module~\cite{philion2020lift} transforms and fuses image features from multiple camera images to unified voxel features. We use the vanilla 3D-ResNet18 as the Voxel Encoder and use 3D-FPN as the Voxel Decoder in both the occupancy forecasting head and flow prediction head of the Future State Prediction Module. The prediction module containing stacked residual convolutional blocks orderly encodes historical 3D features, expands channel dimensions according to the future time horizon $N_f$, and produces future 3D features, as shown in~\figref{fig:our_model_pred}. Referring to the setups of PowerBEV \cite{ijcai2023p120}, the numbers of the three types of residual convolutional blocks in the prediction module are set to 2, 1, and 2, with the kernel size of (3, 3, 1). 

To extend our occupancy forecasting model to 3D instance prediction, our OCFNet predicts occupancy and 3D flow over $t \in [0, N_f]$, corresponding to 5 continuous estimations specifically in our work. Local maxima are first extracted from the estimated occupancy probabilities at $t=0$ following~\cite{ijcai2023p120}, determining the instances' centers. Then, the instances in the following future frames are associated consecutively with the predicted flow. 

To train our OCFNet using the loss defined in Eq.~(4), we set $\lambda_1=\lambda_3=0.5$ and $\lambda_2=0.05$ to balance the optimization for occupancy forecasting, depth reconstruction, and 3D backward centripetal flow prediction. The total parameter number of our OCFNet is 370\,M, the GFLOPs are 6434, and the training-time GPU memory is 57 GB. We believe that our model can serve as a foundational codebase to facilitate future 4D occupancy forecasting works.

\section{Study on Future Time Horizons}
\label{sec:study_on_future_time_horizon}

We further conduct a study on forecasting performance drops with different future time horizons. Since the occupancy grids of static objects do not change in the future time steps unless ground-truth annotations jitter, here we solely focus on the ability to forecast the future occupancy state of movable objects. In this experiment, we post the performance of OpenOccupancy-C, PowerBEV-3D, and our OCFNet for the first-level task and the second-level task since the baseline SPC fails to forecast the inflated \gmo{} mentioned in \secref{sec:assessment}. As shown in \tabref{tab:time_horizon}, our OCFNet$^{\dag}$ remains the best performance for different time horizons in both tasks. In addition, all the baseline approaches show better performance on Lyft-Level5 than nuScenes as the time period for evaluating on Lyft-Level5 is relatively shorter. The closer the timestamp is to the current moment, the easier it is for all the baselines to forecast the occupancy status. 



\begin{figure}[t]
  \centering
  \includegraphics[width=0.96\linewidth]{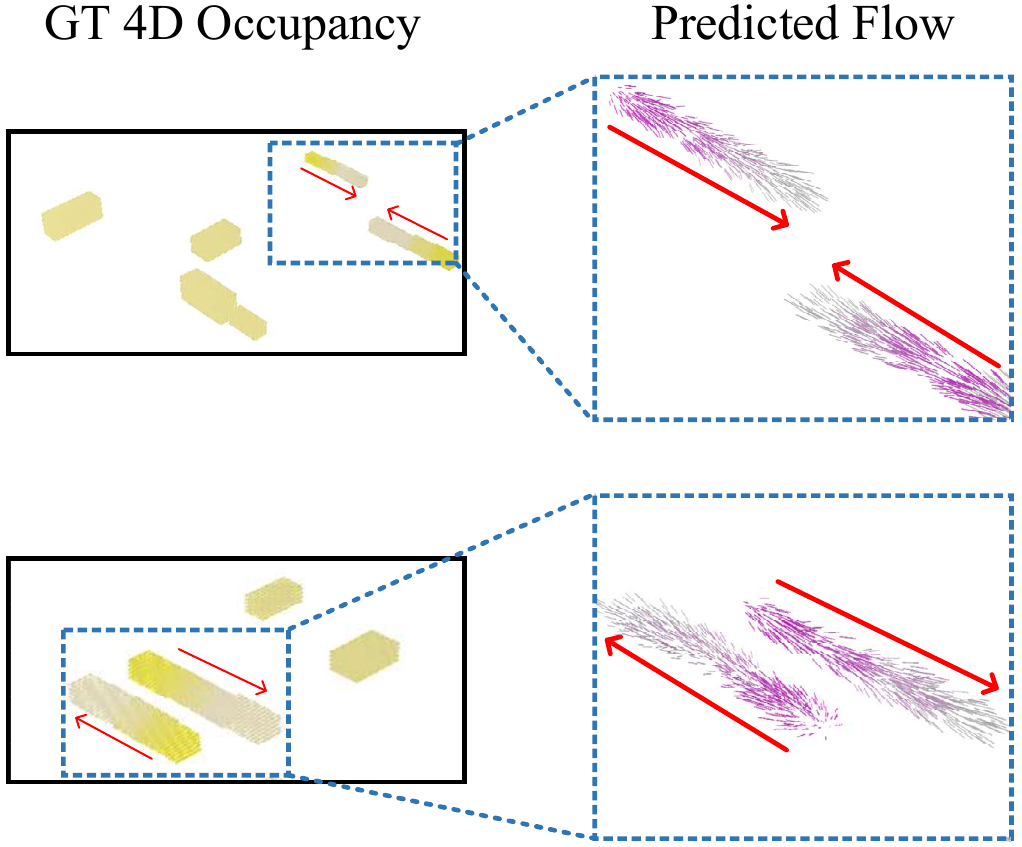}
  \caption{Visualization of predicted 3D backward flow ($t \in [1, N_f]$). The output flow vectors and ground-truth occupancy from timestamps 1 to $N_f$ are assigned colors from dark to light respectively. The motion trend of each selected moving object is represented by red arrows.}
  \label{fig:prediction_flow}
  \vspace{-0.25cm}
\end{figure}

\begin{table*}[t] \small
\setlength{\tabcolsep}{6.2pt}
\center
\renewcommand\arraystretch{1.5}
\caption{Comparison of performance on forecasting \gmo{} in different future time horizons}
\begin{tabular}{lcccc|cccc|cccc}
\toprule
\multicolumn{1}{c}{\multirow{2}{*}{approach}}                & \multicolumn{4}{c|}{nuScenes}     & \multicolumn{4}{c|}{Lyft-Level5}  & \multicolumn{4}{c}{nuScenes-Occupancy} \\ \cline{2-13} 
\multicolumn{1}{c}{}                                                  & 0.5\,s & 1.0\,s & 1.5\,s & 2.0\,s & 0.2\,s & 0.4\,s & 0.6\,s & 0.8\,s & 0.5\,s   & 1.0\,s  & 1.5\,s  & 2.0\,s  \\ \hline
OpenOccupancy-C \cite{wang2023openoccupancy}            & 12.07  & 11.80  & 11.63  & 11.45  & 13.87  & 13.77  & 13.65  & 13.53  & 9.17     & 8.64    & 8.29    & 8.02    \\ \hline
PowerBEV-3D \cite{ijcai2023p120}   & 22.48  & 22.07  & 21.65  & 21.25  & 25.70  & 25.25  & 24.82  & 24.47  & 5.74     & 5.56    & 5.41    & 5.25    \\ \hline
OCFNet (ours)   & 25.95  & 24.92  & 24.33  & 23.89  & 31.51  & 30.87  & 30.17  & 29.56  & 9.17     & 8.72    & 8.53    & 8.35    \\ \cline{1-13} 
OCFNet$^{\dag}$ (ours)    & \textbf{29.36}  & \textbf{28.30}  & \textbf{27.44}  & \textbf{26.82}  &  \textbf{35.58}      & \textbf{34.96}      & \textbf{34.28}       & \textbf{33.56}       &  \textbf{10.64}        & \textbf{10.20}        & \textbf{9.89}        & \textbf{9.68}        \\ \bottomrule
\end{tabular}
\label{tab:time_horizon}
\end{table*}

\section{3D Flow Prediction}
\label{sec:3d_flow_prediction}

Our proposed novel end-to-end occupancy forecasting network OCFNet is trained to reasonably estimate future occupancy state and 3D motion flow simultaneously. We notice that the multi-task learning scheme can help to improve forecasting performance, as shown in \secref{sec:ab_on_flow}. Here, we illustrate the predicted 3D backward centripetal flow in~\figref{fig:prediction_flow}. As can be seen, the predicted flow vectors of the moving object approximately point from the voxel grids of the new coming frame to the ones of the past frame belonging to the same instance. Therefore, the predicted flow can further guide occupancy forecasting by explicitly capturing the motion of GMO in each time interval. Thanks to the flow vectors predicted by \name, we can further associate consistent instances between adjacent future frames, leading to 3D instance prediction beyond occupancy state forecasting.

\begin{figure*}[t]
  \centering
  \includegraphics[width=0.86\linewidth]
  {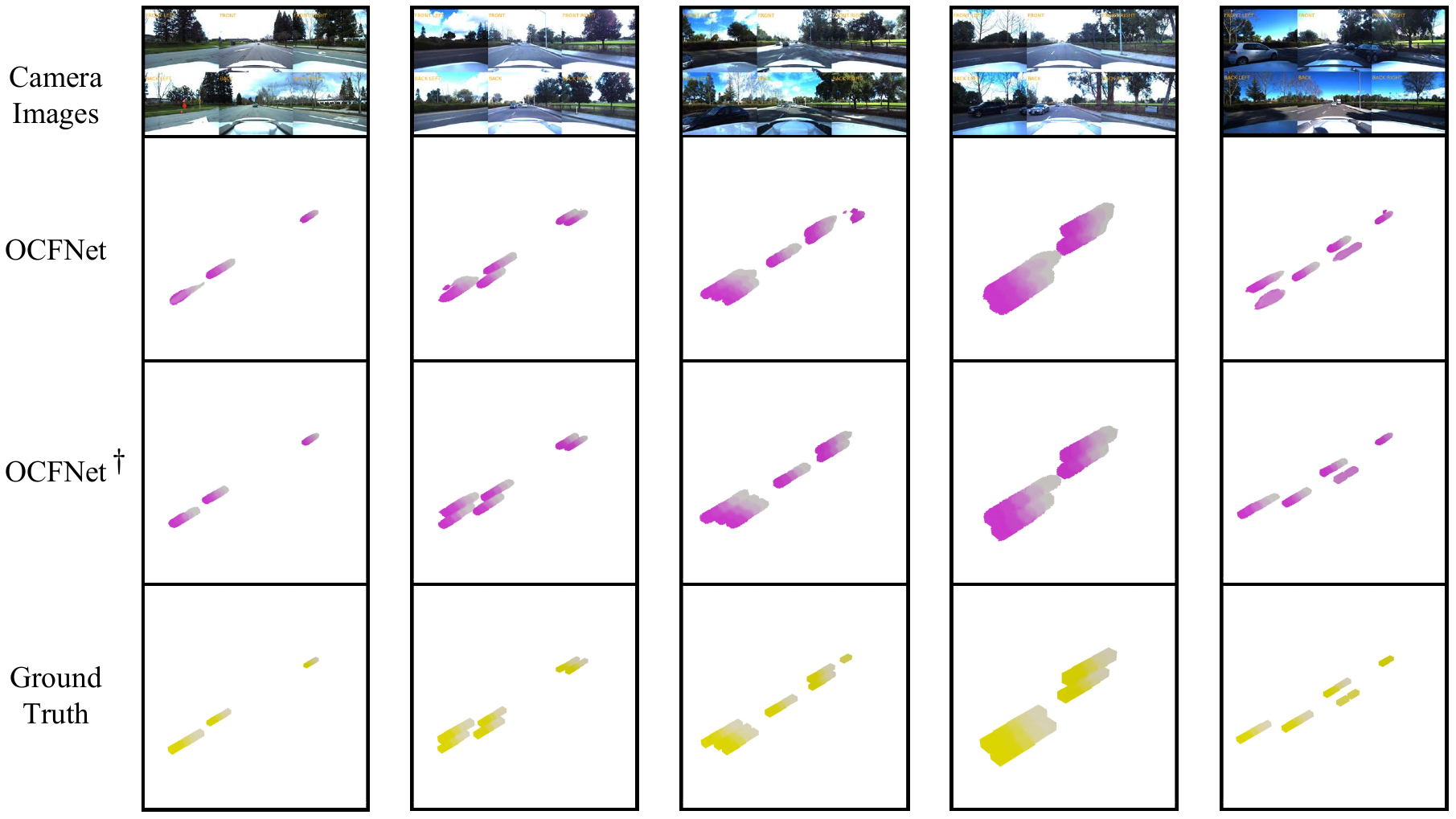}
  \caption{Visualization of forecasting inflated GMO by our proposed OCFNet in small-scale scenes of Lyft-Level5.}
  \label{fig:viz_ocf_small}
\end{figure*}
\begin{figure*}[t]
  \centering
  \includegraphics[width=0.86\linewidth]
  {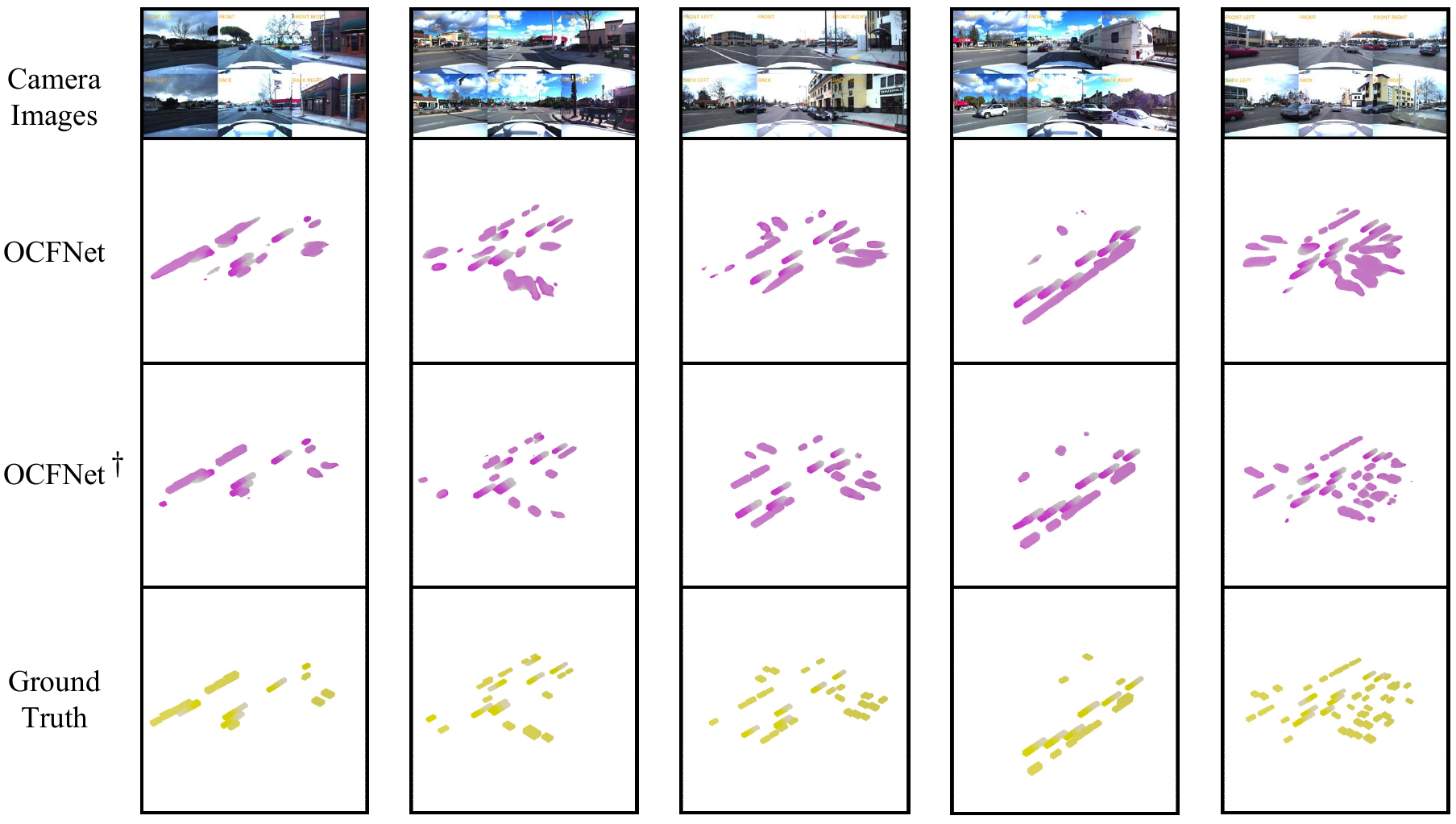}
  \caption{Visualization of forecasting inflated GMO by our proposed OCFNet in large-scale scenes of Lyft-Level5.}
  \label{fig:viz_ocf_large}
\end{figure*}

\vspace{-0.1cm}
\section{3D Instance Prediction}
\label{sec:3d_instance_prediction}

Most existing instance prediction methods \cite{ijcai2023p120, hu2021fiery, casas2021mp3,akan2022stretchbev} can only forecast the future position of objects of interest on BEV representation, while our work extends this task to more complex 3D space. 
We first extract the centers of instances by non-maximum suppression (NMS) at $t=0$ and then associate pixel-wise instance ID over time $t \in [1, N_f]$ using the predicted 3D backward centripetal flow.
To report the instance prediction quality, we extend the metric video panoptic quality (VPQ)~\cite{kim2020video} from the previous 2D instance prediction \cite{hu2021fiery,ijcai2023p120} to our 3D instance prediction, which is calculated by
\begin{small}
\begin{align}
\text{VPQ}_f(\hat{\mathbf{O}}_{f}^{inst}, \mathbf{O}_{f}^{inst}) &= \frac{1}{N_f}\sum_{t=0}^{N_f} \frac{\sum_{\sss{(p_t,q_t) \in TP_t}}\text{IoU}(p_t,q_t)}{|TP_t|+\frac{1}{2}|FP_t|+\frac{1}{2}|FN_t|},
\label{eq:vpq}
\end{align}
\end{small}%
where $TP_t$, $FP_t$, and $FN_t$ represent true positives, false positives, and false negatives at timestamp $t$. Note that in our work the predicted instance is regarded as one true positive once its IoU is greater than $0.2$ (adaptively chosen according to the level of IoU) and the corresponding instance ID is correctly tracked. The experimental results are shown in~\tabref{tab:compare_vpq}. Note that the instance prediction results of PowerBEV-3D are also from the duplication of forecasted 2D flow along the height dimension (same as its 3D extension of forecasted occupancy introduced in \secref{sec:baselines}). As can be seen, our proposed OCFNet$^{\dag}$ shows better 3D instance prediction ability than PowerBEV-3D on Lyft-Level5 while PowerBEV-3D outperforms our approach on nuScenes. In addition, OCFNet$^{\dag}$ improves the prediction of OCFNet by 30.2\% and 13.7\% on nuScenes and Lyft-Level5 respectively. The 2D-3D instance-based prediction baseline presents good instance prediction ability on nuScenes because 2D backward centripetal flow is easier to forecast than the 3D counterpart. On the contrary, our proposed method produces better forecasting results on Lyft-Level5, dominated by far better GMO occupancy forecasting quality of OCFNet$^{\dag}$ than that of PowerBEV-3D on this dataset. Therefore, in the 3D instance prediction task, we further propose a new baseline namely OCFNet$^{*}$, which combines the advantages of our original OCFNet$^{\dag}$ and PowerBEV-3D. The principle is that the 3D flow of the intersection GMO occupancy forecasted by the two methods follows PowerBEV-3D's results, while the other GMO occupancy grids forecasted by OCFNet$^{\dag}$ have the motion flow generated by OCFNet$^{\dag}$ itself. Based on this setup, whether an occupancy grid is occupied totally depends on OCFNet$^{\dag}$, and its flow depends on the choice between OCFNet$^{\dag}$ and PowerBEV-3D. From~\tabref{tab:compare_vpq}, we can see that OCFNet$^{*}$ has the best 3D instance prediction performance, which enhances PowerBEV-3D by 6.7\% on nuScenes and improves OCFNet$^{\dag}$ by 2.1\% on Lyft-Level5.

\begin{table}[t] 
\small
\setlength{\tabcolsep}{14pt}
\center
\renewcommand\arraystretch{1.1}
\caption{Comparison of performance on 3D instance prediction}
\begin{tabular}{l|cc}
\toprule
approaches  & nuScenes             & Lyft-Level5          \\ \cmidrule{1-3}
PowerBEV-3D \cite{ijcai2023p120} & 20.02                     & 27.39                     \\ \cmidrule{1-3}
OCFNet      & 14.26                     & 24.82                     \\
OCFNet$^{\dag}$      & 18.57 & 28.23 \\
OCFNet$^{*}$      & \textbf{21.36} & \textbf{28.81} \\ \bottomrule
\end{tabular}
\label{tab:compare_vpq}
\end{table}

\section{Visualization of future GMO occupancy forecasted by OCFNet on Lyft-Level5}
\label{sec:viz_on_lyft}

In this section, we present our proposed OCFNet forecasting inflated general movable objects of the Lyft-Level5 dataset. \figref{fig:viz_ocf_small} and \figref{fig:viz_ocf_large} show the results in small-scale and large-scale scenes respectively. The prediction results and ground truth from timestamps 1 to $N_f$ are assigned colors from dark to light. As to the small-scale scenes, the valid GMO over the future time horizon occupy relatively fewer volumes and both OCFNet and OCFNet$^{\dag}$ can capture their motion accurately. When it comes to the large-scale conditions, OCFNet$^{\dag}$ significantly outperforms OCFNet which only uses $\frac{1}{6}$ sequences for training. Therefore, when the driving scenario of the ego vehicle has few movable obstacles, such as in rural areas, OCFNet trained with limited data is enough to forecast the future occupancy of surrounding traffic participators. This can significantly improve the deployment efficiency of forecasting modules in autonomous driving systems by decreasing memory consumption and training period.

%




\end{document}